%% file: neurips_2026.tex
\documentclass{article}
\PassOptionsToPackage{numbers,compress}{natbib}
\usepackage[final]{neurips_2026}
\usepackage[utf8]{inputenc}
\usepackage[T1]{fontenc}
\usepackage{hyperref}
\usepackage{url}
\usepackage{booktabs}
\usepackage{multirow}
\usepackage{array}
\usepackage{amsmath}
\usepackage{amssymb}
\usepackage{amsthm}
\usepackage{amsfonts}
\usepackage{nicefrac}
\usepackage{microtype}
\usepackage{xcolor}
\usepackage{graphicx}
\usepackage{subcaption}
\usepackage[inline]{enumitem}
\usepackage{placeins}

\usepackage{makecell}

\makeatletter
\def\@trackname{} 
\makeatother

\title{Preserving Foundational Capabilities in Flow-Matching VLAs through Conservative SFT}

\author{%
  Shanghai Artificial Intelligence Laboratory \\[0.3cm]
  Tianyi Zhang, \quad
  Shaopeng Zhai\thanks{Corresponding author: Shaopeng Zhai (\texttt{zhaishaopeng@pjlab.org.cn}).}, \quad
  Haoran Zhang, \quad
  Fuxian Huang, \quad
  Qi Zhang \\[0.3cm]
  \texttt{\{zhangtianyi, zhaishaopeng\}@pjlab.org.cn} \\[0.3cm]
  \url{https://tyzhang2907.github.io/ConservativeSFT/}
}

\begin{document}

\maketitle

\begin{abstract}
Unconstrained fine-tuning of flow-matching Vision-Language-Action (VLA) models drives dense parameter overwrites, degrading pre-trained capabilities. We present Conservative Supervised Fine-Tuning (ConSFT), an optimization objective that adapts to target distributions while mitigating catastrophic forgetting, requiring zero prior data or architectural overhead. By dynamically scaling learning signals based on model confidence, ConSFT suppresses excessive gradients from low-confidence samples to prevent disproportionate parameter updates, thereby bounding the intrinsic parameter disruption risk. Inspired by reinforcement learning's trust-region clipping, this formulation establishes a progressive learning dynamic to secure target convergence and prior capability retention, maintaining sparse parameter updates without relying on the parallel reference networks required by explicit regularization. We evaluate ConSFT on the LIBERO and RoboTwin benchmarks across state-of-the-art flow-matching VLAs ($\pi_0$, $\pi_{0.5}$, and GR00T-N1.6-3B). The method outperforms vanilla SFT in capability retention by an average absolute margin of over 20\%, matching the efficacy of data-heavy Experience Replay in a prior-data-free regime. Real-world robotic deployments confirm that ConSFT precludes spatial overfitting during downstream adaptation, preserving pre-trained physical skills while acquiring sequential target tasks.
\end{abstract}

\input{Chapters/1_introduction.tex}

\input{Chapters/2_related_work.tex}

\input{Chapters/3_empirical_analysis.tex}

\input{Chapters/4_method.tex}

\input{Chapters/5_experiments.tex}

\input{Chapters/6_conclusion.tex}

\newpage
\bibliographystyle{unsrtnat}
\bibliography{references}

\newpage
\input{Appendix/appendix_main.tex}

\end{document}

%% file: Chapters/1_introduction.tex
\section{Introduction}
\label{sec:introduction}

Large-scale pre-training establishes the foundational capabilities of Vision-Language-Action (VLA) models \cite{wu2026lingbot}. However, deploying these models in novel environments necessitates downstream adaptation to specific target tasks. While vanilla Supervised Fine-Tuning (SFT) directly fits the policy to target demonstrations, adapting large-capacity networks to narrow distributions systematically degrades pre-trained capabilities. These unconstrained parameter updates erode the agent's general robustness and foundational spatial priors, establishing catastrophic forgetting as a primary bottleneck in VLA post-training \cite{chu2025sft, hancock2025actions, liu2026towards}.

Despite the recent adoption of flow-matching in state-of-the-art VLAs \cite{black2024pi0, black2025pi05, bjorck2025groot}, mitigating catastrophic forgetting within these continuous-time architectures remains an open optimization problem. Recent studies in Large Language Models (LLMs) and Vision-Language Models (VLMs) demonstrate that reinforcement learning (RL) resists forgetting by inducing structurally sparse parameter updates \cite{shenfeld2025rlsrazor, mukherjee2026rlsubnetworks}. This capability retention is driven by bounded optimization dynamics, where trust-region constraints restrict the Kullback-Leibler (KL) divergence between the updated and reference policies \cite{schulman2015trpo, schulman2017ppo}. As our empirical analysis in Section \ref{sec:empirical_analysis} confirms, this mechanism translates directly to flow-matching VLAs: unconstrained vanilla SFT drives dense parameter overwrites, whereas bounded updates enforce parameter sparsity and preserve prior capabilities. However, directly applying explicit RL boundary constraints to flow models introduces prohibitive computational overhead. Enforcing exact KL limits requires evaluating action likelihoods by solving probability-flow Ordinary Differential Equations (ODEs) and integrating full continuous trajectories across parallel active and reference networks simultaneously. This integration complexity makes explicit RL trust regions unscalable for large-capacity VLA post-training.

To bypass this integration overhead, we propose Conservative Supervised Fine-Tuning (ConSFT), an optimization objective designed to mitigate catastrophic forgetting in flow-matching VLAs. Inspired by the structural sparsity of RL trust regions, ConSFT emulates these bounded optimization dynamics without requiring prior pre-training data or parallel reference models. Specifically, we dynamically modulate the regression objective based on per-sample model confidence. By down-weighting the learning signals from poorly modeled, low-confidence transitions, the method suppresses unbounded optimization steps. This dynamic weighting establishes a progressive learning dynamic governed by a temperature parameter: the network naturally prioritizes structurally aligned samples before assimilating more complex transitions. Analytically, this formulation enforces a strict bound on the parameter disruption risk per update, preventing the dense global overwrites typical of vanilla SFT. Empirical evaluations across simulated and physical robotic deployments demonstrate that ConSFT effectively retains pre-trained capabilities. In real-world manipulation tasks, our method preserves these foundational capabilities by an absolute success margin of about 20\% against vanilla SFT, without compromising target-task performance.

Our main contributions are summarized as follows:
\begin{itemize}[leftmargin=*]
    \item We identify that catastrophic forgetting in flow-matching VLAs correlates with dense parameter overwrites, providing empirical evidence that RL trust-region mechanisms induce structurally sparse updates and preserve foundational capabilities.
    \item We propose Conservative Supervised Fine-Tuning (ConSFT), an optimization objective inspired by RL's trust-region dynamics that emulates bounded optimization without requiring prior pre-training data or parallel reference models. By dynamically modulating the training objective based on per-sample confidence, ConSFT down-weights learning signals from poorly modeled, low-confidence transitions to suppress unbounded optimization steps.
    \item We mathematically formalize ConSFT as a progressive learning dynamic governed by a temperature parameter. By quantifying the parameter disruption risk via the Fisher Information Matrix (FIM), we analytically prove that this mechanism establishes a strict bound on weight deviations, ensuring updates align with the target distribution without eroding foundational capabilities.
    \item Empirical evaluations across simulated and physical robotic deployments demonstrate that ConSFT preserves pre-trained capabilities by an absolute success margin of about 20\% against vanilla SFT, without compromising target-task performance.
\end{itemize}

%% file: Chapters/2_related_work.tex
\section{Related work}
\label{sec:related_work}

\subsection{RL outperforms SFT in capability retention}
The capability retention discrepancy between Supervised Fine-Tuning (SFT) and Reinforcement Learning (RL) stems from their distinct optimization dynamics \cite{chu2025sft}. SFT minimizes its target objective without constraints, yielding dense parameter updates. This unconstrained fitting induces gradient interference \cite{imanov2026mechanistic} and catastrophic forgetting of pre-trained capabilities \cite{bohnet2025comparative}. In contrast, RL constrains policy updates via trust-region mechanisms, biasing the optimization toward proximal policy distributions \cite{shenfeld2025rlsrazor}. To maximize expected returns within this trust-region constraint, RL parameter updates converge to sparse subnetworks \cite{mukherjee2026rlsubnetworks}. This parameter sparsity explains how bounded optimization protects foundational capabilities from parameter disruption during downstream adaptation.

\subsection{Existing mitigations for catastrophic forgetting}
Methods to mitigate catastrophic forgetting include data replay, policy regularization, and parameter-efficient fine-tuning \cite{kirkpatrick2017overcoming, li2017learning, shi2025continual, luo2025empirical}. Experience Replay (ER) interleaves prior training data during fine-tuning \cite{rolnick2019experience, liu2026pretrained}. For Vision-Language-Action (VLA) models, ER requires access to pre-training corpora, which are proprietary \cite{hu2026simplerecipe}. Regularization methods constrain policy shifts by penalizing divergence from a reference model \cite{ouyang2022training, sun2026safety, phan2025beyond}, an approach extended to flow-matching policies via continuous-time formulations \cite{zhang2026reinflow}. Evaluating this divergence requires a parallel reference network, doubling memory consumption and increasing computational overhead. Low-Rank Adaptation (LoRA) restricts parameter updates to low-rank subspaces to prevent parameter disruption \cite{biderman2024lora, bohnet2025comparative}. This restriction preserves pre-trained capabilities, but the limited capacity reduces plasticity for acquiring downstream skills \cite{biderman2024lora}.

\subsection{Adapting importance sampling ratios for flow-matching models}
Trust region methods like TRPO \cite{schulman2015trpo} and PPO \cite{schulman2017ppo} induce sparse parameter updates to mitigate catastrophic forgetting \cite{mukherjee2026rlsubnetworks, sabbaghi2026robust}, a property recently validated in continuous learning for Vision-Language-Action (VLA) models \cite{lai2025rft, hu2026simplerecipe, liu2026towards}. However, bounding mechanisms such as PPO's clipping and SAPO's soft gating \cite{gao2025sapo} rely on computing exact importance sampling ratios ($\frac{\pi_{\theta}}{\pi_{\text{behavior}}}$). 

This requirement introduces a severe computational bottleneck for modern flow-matching VLA models \cite{black2024pi0, black2025pi05, bjorck2025groot}. Unlike standard autoregressive policies, continuous-time flow-matching architectures lack closed-form action likelihoods ($\log \pi$). Evaluating these probability ratios strictly requires solving probability-flow Ordinary Differential Equations (ODEs), rendering direct computation prohibitively expensive during large-scale fine-tuning.

To bypass this computational overhead, recent flow policy gradient methods \cite{mcallister2025fpo} approximate the flow-matching policy ratio via the exponential loss difference:
\begin{equation}
\frac{\pi_{\theta}(a|s)}{\pi_{\text{behavior}}(a|s)} \approx \exp \Big( \mathcal{L}_{\text{flow}}(\theta_{\text{behavior}}) - \mathcal{L}_{\text{flow}}(\theta) \Big)
\label{eq:flow_policy_ratio}
\end{equation}

where $\mathcal{L}_{\text{flow}}$ denotes the standard continuous-time regression objective. While existing studies explore applying online RL directly to flow-matching models \cite{zhang2026reinflow, pfrommer2025reinforcement}, we build upon this exponential approximation to formulate ConSFT, achieving bounded, sparse parameter updates entirely within the vanilla SFT paradigm.

%% file: Chapters/3_empirical_analysis.tex
\section{Optimization dynamics: trust regions and parameter sparsity}
\label{sec:empirical_analysis}

\subsection{SFT as unconstrained policy optimization}
To understand how RL mechanisms mitigate catastrophic forgetting compared to SFT, we cast both paradigms into a unified policy optimization framework \cite{chu2025sft}. Policy gradients maximize expected returns via the objective $\mathbb{E}_{(s,a) \sim \pi_{\theta}} [\log \pi_{\theta}(a|s) A]$, where the update magnitude is scaled by the advantage $A$. When adapting this objective to off-policy rollouts or offline datasets, the sampling distribution shifts to the target data distribution $\mathcal{D}$. To account for this shift, trust-region methods \cite{schulman2015trpo} introduce the importance sampling ratio $r(\theta) = \frac{\pi_{\theta}(a|s)}{\pi_{\text{behavior}}(a|s)}$. To prevent unconstrained parameter updates, Proximal Policy Optimization (PPO) \cite{schulman2017ppo} constrains this ratio using a trust-region clipping mechanism:
\begin{equation}
\mathcal{L}_{\text{PPO}}(\theta) = - \mathbb{E}_{(s,a) \sim \mathcal{D}} \left[ \min \Big( r(\theta)A, \mathrm{clip}\big(r(\theta), 1-\epsilon, 1+\epsilon\big)A \Big) \right].
\label{eq:ppo_objective}
\end{equation}

SFT reduces to a specific instance of this framework. By fitting the target dataset $\mathcal{D}$ via the unconstrained loss $\mathcal{L}_{\text{flow}}$ \cite{black2024pi0, bjorck2025groot}, SFT assumes all demonstrations are optimal, assigning a constant advantage ($A=1$) to all samples. Because SFT maximizes the likelihood of actions from $\mathcal{D}$ without anchoring to a prior behavior policy, it treats the reference distribution as a uniform prior ($\pi_{\text{behavior}} \to 1$). This optimization eliminates the importance sampling ratio and the associated clipping constraints.

The optimization difference between RL and SFT isolates to two gradient modulators: (1) \textbf{the advantage scalar ($A$)}, which scales updates based on sample quality, and (2) \textbf{trust-region clipping}, which constrains parameter divergence from the reference policy. Isolating these components in an empirical ablation identifies the primary driver of capability retention. Configurations for the PPO baselines, including distributed rollout architectures and reward formulations, are in Appendix \ref{app:subsec:ppo_hyperparameters}.

\subsection{Trust regions mitigate capability degradation}
\label{subsec:forgetting_analysis}

To empirically isolate the mechanisms governing capability retention, we fine-tune a pre-trained $\pi_0$ base model \cite{black2024pi0} on the worst-performing sub-task within the LIBERO-Spatial suite, and measure performance degradation on the held-out Object and Goal suites. To establish a fair benchmark for evaluating forgetting, we train all baselines until they achieve a comparable success rate (87\% to 95\%) on the target task prior to evaluation.

The results summarized in Table \ref{tab:forgetting} reveal a fundamental failure mode in unconstrained optimization. Despite utilizing advantage-weighted gradients to prioritize high-quality actions, PPO-NoClip suffers capability degradation down to a 3\% average prior task performance, characteristic of the vanilla SFT paradigm (9\%) \cite{luo2025empirical}. In contrast, PPO with a clipping boundary preserves an average foundational capability of 39\%, limiting the degradation on the Object suite to 13\% compared to the 57\% drop observed without clipping. To examine this degradation, Appendix \ref{app:subsec:per_task_success} details the per-task capability retention trajectories (Figures \ref{fig:app_object_curves} and \ref{fig:app_goal_curves}). While recent studies \cite{shenfeld2025rlsrazor, mukherjee2026rlsubnetworks, chu2025sft} observe that policy optimization preserves pre-trained capabilities better than vanilla SFT, the underlying mechanism remains debated. Our empirical ablation isolates the explicit trust-region constraint, rather than advantage weighting or on-policy sampling alone, as the indispensable stabilizer in flow-matching action spaces. Mechanistically, when a policy update for a given transition pushes the ratio $r(\theta)$ beyond the $[1-\epsilon, 1+\epsilon]$ boundary, the clipping mechanism truncates its gradient contribution. This operation transforms the objective from unconstrained data fitting to conditional adaptation; the clip acts as a hard constraint that nullifies updates requiring excessive deviation from the pre-trained parameter space. This mathematical boundary prevents dense, unconstrained parameter shifts.

\begin{table*}[t]
\caption{\textbf{Effect of trust regions on capability retention.} While unconstrained methods (vanilla SFT and PPO-NoClip) exhibit severe performance degradation on held-out tasks, PPO with clipping boundaries preserves foundational capabilities. Parentheses show absolute drops ($\downarrow$) from base models.}
\label{tab:forgetting}
\centering
\resizebox{\textwidth}{!}{
\begin{tabular}{l c c c c}
\toprule
\multirow{2}{*}[-0.8ex]{\textbf{Method}} & \textbf{Training (Tgt)} & \multicolumn{3}{c}{\textbf{Prior Task Performance}} \\
\cmidrule(lr){2-2} \cmidrule(lr){3-5}
& \textbf{LIBERO-Spatial} & \textbf{LIBERO-Object} & \textbf{LIBERO-Goal} & \textbf{Average} \\
\midrule
$\pi_0$ Base & 0.25 $\pm$ 0.097 & 0.58 $\pm$ 0.035 & 0.40 $\pm$ 0.035 & 0.49 $\pm$ 0.025 \\
$\pi_0$ SFT & 0.90 $\pm$ 0.067 & 0.02 $\pm$ 0.010 ($\downarrow$ 0.56) & 0.16 $\pm$ 0.026 ($\downarrow$ 0.24) & 0.09 $\pm$ 0.014 ($\downarrow$ 0.40) \\
$\pi_0$ PPO-NoClip & 0.87 $\pm$ 0.075 & 0.01 $\pm$ 0.007 ($\downarrow$ 0.57) & 0.05 $\pm$ 0.015 ($\downarrow$ 0.35) & 0.03 $\pm$ 0.009 ($\downarrow$ 0.46) \\
$\pi_0$ PPO ($\epsilon=0.2$) & 0.95 $\pm$ 0.049 & \textbf{0.45 $\pm$ 0.035 ($\downarrow$ 0.13)} & \textbf{0.32 $\pm$ 0.033 ($\downarrow$ 0.08)} & \textbf{0.39 $\pm$ 0.024 ($\downarrow$ 0.10)} \\
\bottomrule
\end{tabular}
}
\end{table*}

\subsection{Trust regions induce parameter update sparsity}
\label{subsec:3_sparsity_analysis}

To quantify the impact of constraining mechanisms on weight divergence, we compute parameter update sparsity using the relative deviation threshold defined in Appendix \ref{app:subsec:sparsity_definition}. Figure \ref{fig:sparsity} demonstrates that trust-region constraints reduce the global update volume by half, restricting modifications to $15\%$ of the parameter space compared to $30\%$ under unconstrained SFT. At the component level, PPO yields $>99\%$ sparsity across core Attention and MLP weights, aligning with observations in policy optimization \cite{mukherjee2026rlsubnetworks}. Prior studies link sparse updates to the alleviation of catastrophic forgetting \cite{imanov2026mechanistic, shenfeld2025rlsrazor, chu2025sft}. Our results demonstrate that constrained RL in flow-matching VLAs replicates this property. The co-occurrence of parameter sparsity and capability retention indicates a correlation between constraining parameter updates and resisting the degradation of foundational skills.

\begin{figure}[htbp]
    \centering
    \includegraphics[width=\linewidth]{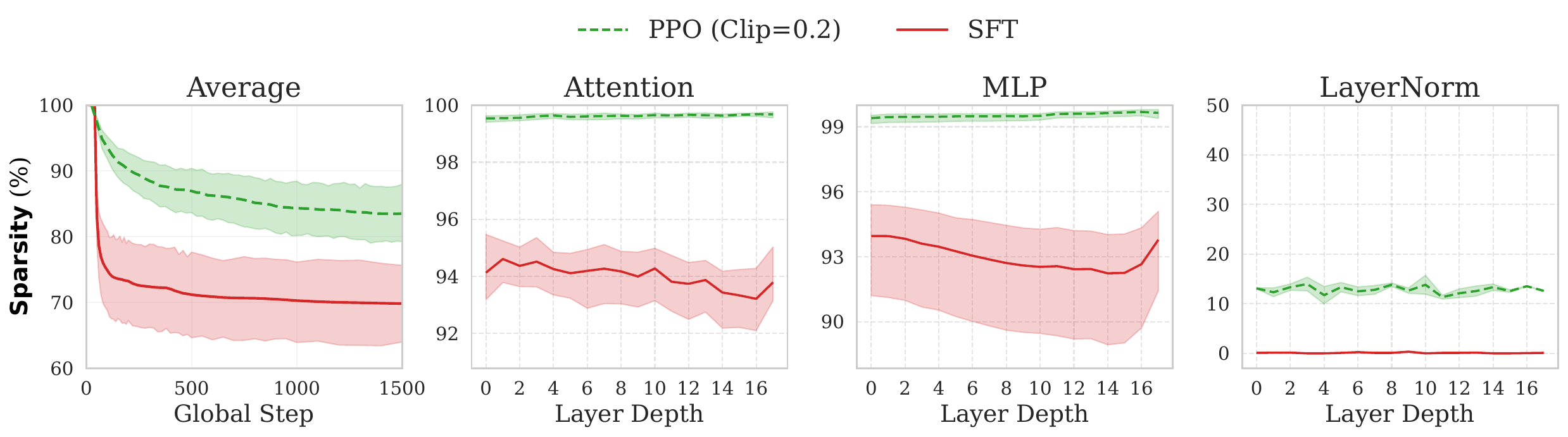}
    \caption{\textbf{Parameter update sparsity across optimization objectives.} \textbf{(Left)} Global sparsity progression. Trust-region constraints (PPO) reduce the update scope compared to unconstrained SFT. \textbf{(Right)} Layer-wise sparsity profiles. PPO yields $>99\%$ sparsity in core Attention and MLP weights.}
    \label{fig:sparsity}
\end{figure}

%% file: Chapters/4_method.tex
\section{Conservative supervised fine-tuning (ConSFT)}
\label{sec:methodology}

We propose Conservative Supervised Fine-Tuning (\textbf{ConSFT}), an optimization objective that emulates the structural sparsity of trust-region methods without the computational overhead of online policy evaluation. By dynamically down-weighting the learning signals of low-confidence transitions via an exponential loss weight, ConSFT analytically bounds the local parameter disruption risk. This formulation restricts weight divergence to highly localized subspaces, enforcing conservative updates entirely within the standard supervised regression paradigm.

\subsection{Conservative weighting via oracle policy substitution}
\label{subsec:conservative_weighting}

In trust-region methods, parameter updates are bounded by the importance sampling ratio $r(\theta) = \pi_{\theta}(a|s) / \pi_{\mathrm{behavior}}(a|s)$. For continuous-time generative models with intractable likelihoods, this ratio is approximated by assuming the intractable Kullback-Leibler (KL) divergence gap remains locally invariant. Following Flow Policy Optimization (FPO) \cite{mcallister2025fpo, yi2026flow}, the flow-matching objective $\mathcal{L}_{\mathrm{flow}}$ serves as a tractable proxy for the negative log-likelihood (or ELBO), up to a parameter-independent noise schedule constant $c$. Because $c$ is independent of the policy, it cancels out when approximating the likelihood ratio:
\begin{equation}
r(\theta) \approx \frac{\exp(\mathrm{ELBO}_{\theta})}{\exp(\mathrm{ELBO}_{\mathrm{behavior}})} = \frac{\exp(-\mathcal{L}_{\theta} + c)}{\exp(-\mathcal{L}_{\mathrm{behavior}} + c)} = \exp\left( \mathcal{L}_{\mathrm{behavior}} - \mathcal{L}_{\theta} \right).
\label{eq:rl_ratio}
\end{equation}

The supervised regression paradigm lacks an explicit behavior policy. To apply the trust-region mechanism, we conceptually define the behavior policy as the ideal expert distribution that perfectly generates the offline demonstrations. Consequently, the theoretical flow matching loss of this ideal policy on its own trajectory distribution evaluates to strictly zero ($\mathcal{L}_{\mathrm{behavior}} = 0$). By substituting this property alongside $\mathcal{L}_{\theta} = \mathcal{L}_{\mathrm{SFT}}(\theta)$ into Equation \ref{eq:rl_ratio}, and introducing a temperature parameter $\tau > 0$ to regulate the trust-region boundary, the \textbf{Conservative Importance Weight} evaluates as:
\begin{equation}
\omega_{\mathrm{ConSFT}}(\theta) = \exp\left( \frac{\mathcal{L}_{\mathrm{behavior}} - \mathcal{L}_{\mathrm{SFT}}(\theta)}{\tau} \right) = \exp\left( -\frac{\mathcal{L}_{\mathrm{SFT}}(\theta)}{\tau} \right).
\label{eq:consft_weight}
\end{equation}

To enforce this exponential weight as a dynamic modulation scale rather than a differentiable target, we apply a stop-gradient ($\mathrm{sg}$) operator. The final \textbf{ConSFT objective} is defined as:
\begin{equation}
\mathcal{J}_{\mathrm{ConSFT}}(\theta) = \mathrm{sg}\left[ \exp\left(-\frac{\mathcal{L}_{\mathrm{SFT}}(\theta)}{\tau}\right) \right] \cdot \mathcal{L}_{\mathrm{SFT}}(\theta).
\label{eq:consft_objective}
\end{equation}

This exponentiated weight induces a \textbf{progressive learning dynamic} that addresses a limitation of vanilla SFT: unconstrained fitting of low-confidence (high-loss) transitions generates large gradient updates that disrupt foundational weight configurations \cite{chu2025sft, imanov2026mechanistic}. ConSFT mitigates this failure mode. During early adaptation, it suppresses excessive gradients from unfamiliar samples to prioritize structurally aligned subsets. As the network adapts, its confidence on previously difficult transitions increases over the course of training, amplifying their update weights. This continuous feedback loop ensures an ordered assimilation of the target distribution, preventing dense, unbounded parameter overwrites.

\subsection{ConSFT formulation: A trust-region inspired regression objective}
\label{subsec:risk_analysis}

To mathematically quantify how this progressive learning dynamic protects foundational skills, we evaluate the expected impact of gradient updates on previously acquired capabilities. From a local second-order approximation perspective, a generic positive semi-definite Fisher Information Matrix (FIM), denoted as $F$, captures the curvature of the pre-training objective. Following established formulations \cite{kirkpatrick2017overcoming, lai2025rft}, we adopt the concept of \textit{forgetting risk}, where the expected performance degradation on prior tasks induced by a target-task gradient update $g$ is strictly bounded by its quadratic form:
\begin{equation}
\mathcal{R}(g) = g^{\top} F g.
\end{equation}

In vanilla SFT, the objective assigns uniform static weighting, driving the optimization trajectory via the unconstrained gradient $g_{\mathrm{SFT}} = \nabla_{\theta}\mathcal{L}_{\mathrm{SFT}}(\theta)$. This yields a baseline forgetting risk of $\mathcal{R}(g_{\mathrm{SFT}}) = g_{\mathrm{SFT}}^{\top} F g_{\mathrm{SFT}}$. By unconditionally fitting target task demonstrations, SFT produces gradients that maximize this quadratic risk, directly resulting in catastrophic forgetting. Since the magnitude of the gradient $g_{\mathrm{SFT}}$ positively correlates with the regression loss $\mathcal{L}_{\mathrm{SFT}}(\theta)$, high-loss target samples disproportionately amplify the forgetting risk.

In ConSFT, the gradient is dynamically modulated by the scalar weight $\omega = \exp(-\mathcal{L}_{\mathrm{SFT}}(\theta)/\tau)$. Because this modulation acts purely as a scalar multiplier on the gradient ($g_{\mathrm{ConSFT}} = \omega \cdot g_{\mathrm{SFT}}$), the resulting forgetting risk attenuates quadratically, independent of the specific curvature $F$:
\begin{equation}
\label{eq:risk_attenuation}
\mathcal{R}(g_{\mathrm{ConSFT}}) = \omega^2 \cdot \mathcal{R}(g_{\mathrm{SFT}}) = \exp\left(-\frac{2\mathcal{L}_{\mathrm{SFT}}(\theta)}{\tau}\right) \mathcal{R}(g_{\mathrm{SFT}}).
\end{equation}

Equation \ref{eq:risk_attenuation} reveals the mathematical mechanism of ConSFT. For unfamiliar transitions where $\mathcal{L}_{\mathrm{SFT}}(\theta)$ is large, the exponential coefficient exponentially decays toward zero. This scaling actively counteracts the positive correlation between high losses and large gradient magnitudes, strictly suppressing the intrinsic forgetting risk. This implicit regularization guarantees that the objective does not generate updates that erase foundational skills to accommodate difficult target samples. Conversely, as the model masters specific transitions and $\mathcal{L}_{\mathrm{SFT}}(\theta) \to 0$, the risk profile asymptotically recovers toward the unconstrained SFT baseline, restoring full adaptation capacity.

%% file: Chapters/5_experiments.tex
\section{Experiments}
\label{sec:experiments}

Our empirical evaluation addresses four core questions:
\begin{enumerate}[label=(Q\arabic*), leftmargin=*]
    \item Does ConSFT successfully balance target-task adaptation with the preservation of pre-trained capabilities across distinct flow-matching VLA backbones? (Section \ref{subsec:preserving_capabilities_flow_matching})
    \item Does ConSFT outperform established regularization and replay baselines in capability retention, while circumventing their memory and computational overheads? (Section \ref{subsec:baseline_comparison})
    \item Mechanistically, does the conservative weighting objective confine parameter updates to sparse subspaces? (Section \ref{subsec:5_sparsity_analysis})
    \item Do the capability retention benefits of ConSFT translate to robust multi-task execution during real-world robotic deployments? (Section \ref{subsec:real_world})
\end{enumerate}

\subsection{Experimental setup}
\label{subsec:experimental_setup}

\textbf{Benchmarks \& architectures.} We evaluate ConSFT on two continuous control benchmarks: \textbf{LIBERO} \cite{liu2023libero} (\textit{Spatial}, \textit{Object}, and \textit{Goal} suites) for capability retention, and \textbf{RoboTwin} \cite{mu2025robotwin} for bimanual coordination. The evaluation spans three representative flow-matching VLA architectures: $\pi_0$ \cite{black2024pi0}, $\pi_{0.5}$ \cite{black2025pi05}, and GR00T-N1.6-3B \cite{bjorck2025groot}.

\textbf{Implementation details.} ConSFT operates with zero auxiliary memory overhead, requiring no parallel reference networks. We apply the dynamic confidence weight (Equation \ref{eq:consft_objective}) on a per-sample basis via a stop-gradient operator. The temperature $\tau$ is exponentially annealed during training, asymptotically relaxing the optimization from bounded parameter updates back to unconstrained regression (detailed in Appendix \ref{app:subsec:temperature_schedule}). Distributed training infrastructure and comprehensive hyperparameter configurations are provided in Appendix \ref{app:subsec:vla_hyperparameters}.

\subsection{Capability retention in flow-matching VLAs}
\label{subsec:preserving_capabilities_flow_matching}

Table \ref{tab:main_results} details the downstream adaptation and prior task retention across the LIBERO and RoboTwin benchmarks. Beyond the raw performance metrics, the results highlight two primary optimization dynamics:

\textbf{ConSFT attenuates forgetting risk.} Unconstrained fine-tuning on narrowly defined target distributions rapidly overwrites prior capabilities. For example, fitting the base $\pi_0$ architecture to the LIBERO-Spatial task via vanilla SFT induces severe failures on previously learned suites, degrading the LIBERO-Object success rate from $58\%$ to $2\%$ ($\downarrow 56\%$). In contrast, ConSFT restricts this degradation while ensuring convergence on the target task. For the same $\pi_0$ model, ConSFT limits the LIBERO-Object drop to $\downarrow 26\%$ while matching SFT's exact target-task performance on LIBERO-Spatial ($90\%$). This empirical retention directly validates the analytical risk attenuation established in Equation \ref{eq:risk_attenuation}, demonstrating that the exponential scaling mechanism effectively prevents the objective from generating gradients that disrupt FIM-sensitive foundational weights.

\textbf{Risk suppression prevents collapse under severe distribution shifts.} The RoboTwin benchmark introduces a structural discrepancy between the adaptation target and foundational competencies. Specifically, the model is fine-tuned exclusively on independent dual-arm operations (\textit{RoboTwin-Indep.}), whereas retention is evaluated on held-out single-arm (\textit{RoboTwin-Single-Arm}) and tightly coordinated bimanual tasks (\textit{RoboTwin-Coord.}). Driven by this task discrepancy, vanilla SFT suffers a near-total degradation of previously acquired skills, collapsing to $0\%$ success on RoboTwin-Coord. for both the $\pi_0$ and $\pi_{0.5}$ architectures. By dynamically suppressing the forgetting risk, ConSFT preserves these held-out capabilities, retaining a $13\%$ success rate on RoboTwin-Coord. for $\pi_0$. Crucially, this retention does not compromise target task execution; ConSFT achieves a $60\%$ success rate on RoboTwin-Indep., slightly outperforming the unconstrained SFT baseline ($55\%$). This demonstrates that conservative weighting actively stabilizes the learning trajectory, even when target demonstrations diverge significantly from the foundational distribution.

\begin{table*}[t]
\caption{\textbf{Mitigation of catastrophic forgetting.} ConSFT preserves prior capabilities on LIBERO and RoboTwin. Parentheses show absolute drops ($\downarrow$) from base models.}
\label{tab:main_results}
\centering
\resizebox{\textwidth}{!}{
\begin{tabular}{l l c c c c}
\toprule
\multirow{2}{*}[-0.6ex]{\textbf{Architecture}} & \multirow{2}{*}[-0.6ex]{\textbf{Method}} & \textbf{Target} & \multicolumn{3}{c}{\textbf{Prior Capabilities}} \\
\cmidrule(lr){3-3} \cmidrule(lr){4-6}
& & \textbf{LIBERO-Spatial} & \textbf{LIBERO-Object} & \textbf{LIBERO-Goal} & \textbf{Average} \\
\midrule
\multirow{3}{*}{$\pi_0$} 
& Base & 0.25 $\pm$ 0.097 & 0.58 $\pm$ 0.035 & 0.40 $\pm$ 0.035 & 0.49 $\pm$ 0.025 \\
& SFT & \textbf{0.90 $\pm$ 0.067} & 0.02 $\pm$ 0.010 ($\downarrow$ 0.56) & 0.16 $\pm$ 0.026 ($\downarrow$ 0.24) & 0.09 $\pm$ 0.014 ($\downarrow$ 0.40) \\
& \textbf{ConSFT} & \textbf{0.90 $\pm$ 0.067} & \textbf{0.32 $\pm$ 0.033 ($\downarrow$ 0.26)} & \textbf{0.35 $\pm$ 0.034 ($\downarrow$ 0.05)} & \textbf{0.34 $\pm$ 0.024 ($\downarrow$ 0.15)} \\
\midrule
\multirow{3}{*}{$\pi_{0.5}$} 
& Base & 0.10 $\pm$ 0.067 & 0.80 $\pm$ 0.028 & 0.74 $\pm$ 0.031 & 0.77 $\pm$ 0.021 \\
& SFT & \textbf{1.00 $\pm$ 0.000} & 0.18 $\pm$ 0.027 ($\downarrow$ 0.62) & 0.28 $\pm$ 0.032 ($\downarrow$ 0.46) & 0.23 $\pm$ 0.021 ($\downarrow$ 0.54) \\
& \textbf{ConSFT} & \textbf{1.00 $\pm$ 0.000} & \textbf{0.46 $\pm$ 0.035 ($\downarrow$ 0.34)} & \textbf{0.40 $\pm$ 0.035 ($\downarrow$ 0.34)} & \textbf{0.43 $\pm$ 0.025 ($\downarrow$ 0.34)} \\
\midrule
\multirow{3}{*}{GR00T} 
& Base & 0.45 $\pm$ 0.111 & 0.88 $\pm$ 0.023 & 0.90 $\pm$ 0.021 & 0.89 $\pm$ 0.016 \\
& SFT & \textbf{0.70 $\pm$ 0.102} & 0.42 $\pm$ 0.035 ($\downarrow$ 0.46) & 0.55 $\pm$ 0.035 ($\downarrow$ 0.35) & 0.49 $\pm$ 0.025 ($\downarrow$ 0.40) \\
& \textbf{ConSFT} & 0.63 $\pm$ 0.108 & \textbf{0.56 $\pm$ 0.035 ($\downarrow$ 0.32)} & \textbf{0.62 $\pm$ 0.034 ($\downarrow$ 0.28)} & \textbf{0.59 $\pm$ 0.025 ($\downarrow$ 0.30)} \\
\midrule
& & \textbf{RoboTwin-Indep.} & \textbf{RoboTwin-Single-Arm} & \textbf{RoboTwin-Coord.} & \textbf{Average} \\
\midrule
\multirow{3}{*}{$\pi_0$} 
& Base & 0.10 $\pm$ 0.067 & 0.57 $\pm$ 0.042 & 0.30 $\pm$ 0.072 & 0.44 $\pm$ 0.037 \\
& SFT & 0.55 $\pm$ 0.111 & 0.27 $\pm$ 0.038 ($\downarrow$ 0.30) & 0.00 $\pm$ 0.000 ($\downarrow$ 0.30) & 0.14 $\pm$ 0.026 ($\downarrow$ 0.30) \\
& \textbf{ConSFT} & \textbf{0.60 $\pm$ 0.110} & \textbf{0.43 $\pm$ 0.042 ($\downarrow$ 0.14)} & \textbf{0.13 $\pm$ 0.053 ($\downarrow$ 0.17)} & \textbf{0.28 $\pm$ 0.033 ($\downarrow$ 0.16)} \\
\midrule
\multirow{3}{*}{$\pi_{0.5}$} 
& Base & 0.25 $\pm$ 0.097 & 0.57 $\pm$ 0.042 & 0.25 $\pm$ 0.068 & 0.41 $\pm$ 0.037 \\
& SFT & \textbf{0.65 $\pm$ 0.107} & 0.25 $\pm$ 0.037 ($\downarrow$ 0.32) & 0.00 $\pm$ 0.000 ($\downarrow$ 0.25) & 0.13 $\pm$ 0.025 ($\downarrow$ 0.28) \\
& \textbf{ConSFT} & \textbf{0.65 $\pm$ 0.107} & \textbf{0.40 $\pm$ 0.041 ($\downarrow$ 0.17)} & \textbf{0.10 $\pm$ 0.047 ($\downarrow$ 0.15)} & \textbf{0.25 $\pm$ 0.032 ($\downarrow$ 0.16)} \\
\bottomrule
\end{tabular}
}
\end{table*}

\subsection{Mitigating forgetting without data replay or structural constraints}
\label{subsec:baseline_comparison}

To benchmark ConSFT against established anti-forgetting paradigms, we evaluate three methods adapted for flow-matching VLAs: KL-Regularization (LwF) \cite{li2017learning}, Experience Replay (ER) \cite{rolnick2019experience}, and Low-Rank Adaptation (LoRA) \cite{biderman2024lora, bohnet2025comparative} (implementation details in Appendix \ref{app:subsec:baseline_details}). Table \ref{tab:baselines_comparison} summarizes the aggregate retention performance.

Explicit KL constraints on continuous vector fields (LwF) provide insufficient protection against capability degradation, with success rates collapsing to $0\%$ on both LIBERO-Goal and RoboTwin-Coord. Concurrently, LwF restricts target-task convergence, achieving only a $45\%$ success rate on RoboTwin-Indep. LoRA restricts parameter updates to low-rank subspaces, yielding retention on specific suites (e.g., $\downarrow 2\%$ on LIBERO-Goal). However, this structural constraint provides inconsistent preservation, degrading to $10\%$ on LIBERO-Object, and restricts the model's adaptation capacity on target tasks, evidenced by a $51\%$ success rate on RoboTwin-Indep. Experience Replay (ER) interleaves historical data to mitigate degradation; however, despite this explicit data dependency, ER achieves lower average retention across both LIBERO ($18\%$) and RoboTwin ($15\%$) compared to our method.

By directly modulating the update scale via conservative weighting, ConSFT bypasses the need for external data buffers or low-rank structural bottlenecks. It achieves the highest average prior capability retention across both domains ($34\%$ and $28\%$) while matching or exceeding the target-task performance of all baselines (e.g., $90\%$ on LIBERO-Spatial and $60\%$ on RoboTwin-Indep.). To examine these optimization dynamics temporally, Appendix \ref{app:subsec:consft_per_task_success} details the per-task retention trajectories (Figures \ref{fig:app_object_consft} and \ref{fig:app_goal_consft}), confirming that ConSFT attenuates the forgetting risk inherent to the gradient update.

\begin{table*}[t]
\caption{\textbf{Capability retention across continual learning baselines.} ConSFT mitigates catastrophic forgetting without requiring prior data replay or structural constraints. Values in parentheses denote the absolute performance drop ($\downarrow$) compared to the base model ($\pi_0$).}
\label{tab:baselines_comparison}
\centering
\resizebox{\textwidth}{!}{
\begin{tabular}{l c c c c}
\toprule
\multirow{2}{*}[-0.8ex]{\textbf{Method}} & \textbf{Target} & \multicolumn{3}{c}{\textbf{Prior Capabilities}} \\
\cmidrule(lr){2-2} \cmidrule(lr){3-5}
& \textbf{LIBERO-Spatial} & \textbf{LIBERO-Object} & \textbf{LIBERO-Goal} & \textbf{Average} \\
\midrule
Base Model & 0.25 $\pm$ 0.097 & 0.58 $\pm$ 0.035 & 0.40 $\pm$ 0.035 & 0.49 $\pm$ 0.025 \\
\midrule
LwF & 0.80 $\pm$ 0.089 & 0.10 $\pm$ 0.021 ($\downarrow$ 0.48) & 0.00 $\pm$ 0.000 ($\downarrow$ 0.40) & 0.05 $\pm$ 0.011 ($\downarrow$ 0.44) \\
ER & \textbf{0.90 $\pm$ 0.067} & 0.20 $\pm$ 0.028 ($\downarrow$ 0.38) & 0.16 $\pm$ 0.026 ($\downarrow$ 0.24) & 0.18 $\pm$ 0.019 ($\downarrow$ 0.31) \\
LoRA & \textbf{0.90 $\pm$ 0.067} & 0.10 $\pm$ 0.021 ($\downarrow$ 0.48) & \textbf{0.38 $\pm$ 0.034 ($\downarrow$ 0.02)} & 0.24 $\pm$ 0.021 ($\downarrow$ 0.25) \\
\textbf{ConSFT} & \textbf{0.90 $\pm$ 0.067} & \textbf{0.32 $\pm$ 0.033 ($\downarrow$ 0.26)} & 0.35 $\pm$ 0.034 ($\downarrow$ 0.05) & \textbf{0.34 $\pm$ 0.024 ($\downarrow$ 0.15)} \\
\midrule
& \textbf{RoboTwin-Indep.} & \textbf{RoboTwin-Single-Arm} & \textbf{RoboTwin-Coord.} & \textbf{Average} \\
\midrule
Base Model & 0.10 $\pm$ 0.067 & 0.57 $\pm$ 0.042 & 0.30 $\pm$ 0.072 & 0.44 $\pm$ 0.037 \\
\midrule
LwF & 0.45 $\pm$ 0.111 & 0.26 $\pm$ 0.037 ($\downarrow$ 0.31) & 0.00 $\pm$ 0.000 ($\downarrow$ 0.30) & 0.13 $\pm$ 0.025 ($\downarrow$ 0.31) \\
ER & \textbf{0.60 $\pm$ 0.110} & 0.30 $\pm$ 0.039 ($\downarrow$ 0.27) & 0.00 $\pm$ 0.000 ($\downarrow$ 0.30) & 0.15 $\pm$ 0.027 ($\downarrow$ 0.29) \\
LoRA & 0.51 $\pm$ 0.112 & 0.35 $\pm$ 0.040 ($\downarrow$ 0.22) & 0.11 $\pm$ 0.049 ($\downarrow$ 0.19) & 0.23 $\pm$ 0.031 ($\downarrow$ 0.21) \\
\textbf{ConSFT} & \textbf{0.60 $\pm$ 0.110} & \textbf{0.43 $\pm$ 0.042 ($\downarrow$ 0.14)} & \textbf{0.13 $\pm$ 0.053 ($\downarrow$ 0.17)} & \textbf{0.28 $\pm$ 0.033 ($\downarrow$ 0.16)} \\
\bottomrule
\end{tabular}
}
\end{table*}

\subsection{Mechanistic validation: conservative weighting induces stable sparsity decay}
\label{subsec:5_sparsity_analysis}

\begin{figure}[htbp]
\centering
\includegraphics[width=\linewidth]{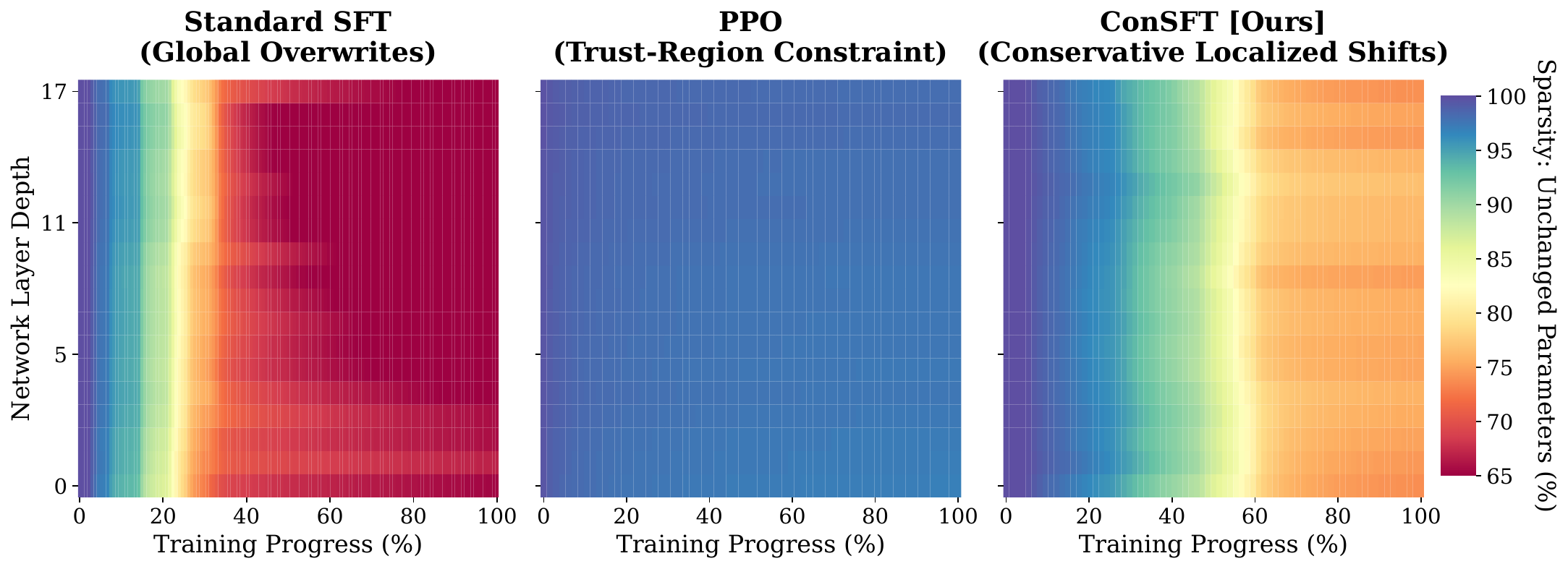}
\caption{\textbf{Evolution of layer-wise update sparsity across training steps.} Vanilla SFT (left) drives a rapid, early collapse in parameter sparsity, resulting in dense global overwrites. ConSFT (right) structurally delays this shift, enforcing a controlled and uniformly decaying optimization trajectory. This progressive adaptation bridges the strict trust-region bounds of PPO (center) and the unconstrained regression of vanilla SFT.}
\label{fig:sparsity_heatmap}
\end{figure}

Figure \ref{fig:sparsity_heatmap} visualizes the temporal evolution of layer-wise parameter updates across the $\pi_0$ architecture. Vanilla SFT (left) triggers rapid, global parameter overwrites early in the training phase. This rapid loss of parameter sparsity directly accounts for the pre-trained capability degradation observed in Section \ref{subsec:preserving_capabilities_flow_matching}.

By bounding the intrinsic disruption risk via the conservative weighting mechanism (Equation \ref{eq:risk_attenuation}), ConSFT fundamentally alters this optimization trajectory. Structural modifications are mathematically inevitable when fitting a completely unfamiliar target distribution, naturally rendering the final converged ConSFT network less sparse than the strictly constrained PPO policy (center) \cite{shenfeld2025rlsrazor, mukherjee2026rlsubnetworks}. However, ConSFT substantially delays the onset of severe sparsity collapse. Rather than permitting immediate global overwrites, the conservative objective enforces a controlled, uniformly decaying sparsity path. This structural modulation mitigates disruptive weight deviations during the critical early phases of adaptation. Consequently, ConSFT achieves an optimal retention-convergence equilibrium: it permits the necessary parameter shifts for target task convergence while strictly bounding the overall disruption, effectively replicating the capability retention properties of on-policy RL without its computational overhead.

\subsection{Real-world deployment}
\label{subsec:real_world}

To evaluate practical deployment, we fine-tune a pre-trained $\pi_{0.5}$ model (capable of object sorting under visual clutter) on a high-precision, long-horizon target task: transferring test tubes between racks. To isolate capability retention, all methods are trained to a controlled 70\% target success rate before being re-evaluated on the prior semantic grasping tasks. Specific configurations for both the target adaptation and prior evaluation tasks are detailed in Appendix \ref{app:subsec:real_world_deployment}. 

\begin{figure}[htbp]
\centering
\includegraphics[width=\linewidth]{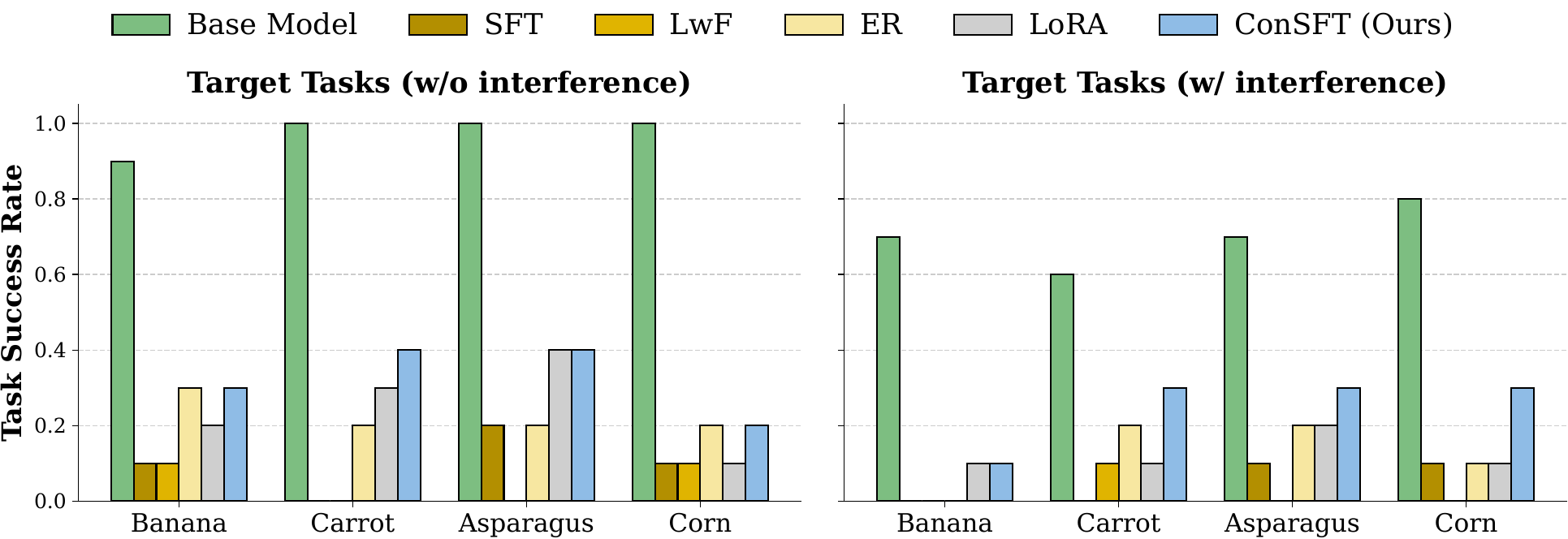}
\caption{\textbf{Capability retention in physical deployments.} Following downstream adaptation to the test-tube target task (controlled at 70\% target success), unconstrained adaptation baselines (vanilla SFT, LwF) exhibit severe degradation of pre-trained capabilities. In contrast, ConSFT achieves the highest prior task retention among all baselines in a prior-data-free regime, maintaining robust performance even under visually cluttered conditions (\textit{w/ interference}).}
\label{fig:real_robot_results}
\end{figure}

\textbf{Case study: Preventing spatial drift.} Trajectory analysis (Figure \ref{fig:real_robot_results}) reveals a critical failure mode in the baselines. Following target task adaptation, vanilla SFT and regularization methods generally retain the visual capacity to locate prior objects. However, their spatial execution degrades severely: the gripper systematically closes prematurely at an elevated z-axis position. This vertical drift stems directly from overfitting to the spatial bias of the target task, which requires high-altitude grasps for tall test tubes. Unconstrained optimization overwrites pre-trained spatial precision to accommodate this shift. By bounding parameter deviations, ConSFT mitigates this spatial drift, securing target task proficiency while preserving foundational spatial capabilities without relying on prior data buffers.

%% file: Chapters/6_conclusion.tex
\section{Conclusion and limitations}
\label{sec:conclusion}

We introduce ConSFT, a prior-data-free optimization framework that bounds capability degradation in flow-matching VLAs. However, this mechanism presents two specific algorithmic limitations. First, by dynamically down-weighting gradients from low-confidence transitions, the conservative objective intrinsically suppresses learning signals from highly unfamiliar target demonstrations. This confidence-based modulation prevents structural disruption but inevitably decelerates target task convergence when acquiring fundamentally new control primitives. Second, while ConSFT strictly restricts parameter-space deviations, it provides no formal theoretical guarantee on the bounded error of the flow-matching ODE trajectories during multi-step inference. Future work will investigate adaptive weighting schedules to accelerate novel skill acquisition and explore trajectory-aware risk constraints to address these mathematical gaps.

%% file: Appendix/appendix_main.tex
\appendix
\input{Appendix/1_appendix_empirical_analysis.tex}
\input{Appendix/2_appendix_method.tex}

\input{Appendix/3_appendix_experiments.tex}

%% file: Appendix/1_appendix_empirical_analysis.tex
\section{Mechanistic ablations: policy optimization and sparsity analysis}
\label{app:sec:rl_details}

This section details the online reinforcement learning configuration for the empirical analysis in Section~\ref{sec:empirical_analysis}. The Proximal Policy Optimization (PPO) baselines initialize from the same pre-trained $\pi_0$ architecture and weights used for supervised fine-tuning.

\subsection{PPO training configuration and reward formulation}
\label{app:subsec:ppo_hyperparameters}

Unlike vanilla SFT, the PPO baseline collects online rollouts via direct environmental interaction. Table~\ref{tab:ppo_hyperparameters} summarizes the hyperparameter configuration, which incorporates two key implementation choices:

First, to mitigate the inference latency of flow-matching ODE integration during online data collection, we deploy an asynchronous distributed training architecture comprising one optimization node and four rollout workers. Second, we adopt a sparse reward function ($r \in \{0, 1\}$). Avoiding dense reward shaping isolates the optimization dynamics and exploration limits of the algorithms from human-engineered task biases.

\begin{table}[htbp]
\caption{\textbf{Hyperparameter configuration for PPO baselines.} Parameters governing the online reinforcement learning evaluations (PPO and PPO-NoClip) for sparse-reward flow-matching VLAs.}
\label{tab:ppo_hyperparameters}
\centering
\begin{tabular}{lc}
\toprule
\textbf{Hyperparameter} & \textbf{Value} \\
\midrule
\multicolumn{2}{l}{\textit{Environment \& Hardware Setup}} \\
Hardware Distribution & 5$\times$ NVIDIA A100 (1 Train, 4 Eval Workers) \\
Reward Structure & Sparse (0 for failure, 1 for success) \\
\midrule
\multicolumn{2}{l}{\textit{Policy \& Rollout Configuration}} \\
ODE Denoising Steps ($K$) & 4 \\
Action Chunking Size & 5 \\
Exploration Noise Level & 0.5 \\
Buffer Capacity & 700 \\
Reuse Threshold & 3 \\
\midrule
\multicolumn{2}{l}{\textit{PPO \& GAE Objective}} \\
Discount Factor ($\gamma$) & 0.99 \\
GAE Parameter ($\lambda$) & 0.95 \\
Trust Region Clip Threshold ($\epsilon$) & 0.2 \\
Entropy Coefficient & 0.0 \\
Critic Warmup Iterations & 0 \\
\midrule
\multicolumn{2}{l}{\textit{Optimization \& Batching}} \\
Optimizer & Adam \\
Adam Betas ($\beta_1, \beta_2$) & (0.9, 0.95) \\
Adam Epsilon ($\epsilon$) & $1.0 \times 10^{-5}$ \\
Actor Learning Rate & $1.0 \times 10^{-6}$ \\
Critic Learning Rate & $1.0 \times 10^{-5}$ \\
Global Batch Size & 1024 \\
Mini-batch Size & 32 \\
Max Gradient Norm & 2.0 \\
\bottomrule
\end{tabular}
\end{table}

\subsection{Detailed per-task retention trajectories}
\label{app:subsec:per_task_success}

To examine capability retention, we track the per-task success rate trajectories for the held-out suites during fine-tuning. We restrict the downstream adaptation to a single target task: \textit{LIBERO-Spatial Task 9}. This environment was selected due to the low initial zero-shot success rate (25\%) of the pre-trained $\pi_0$ base model. Adapting to a narrow target task serves as an optimization stress test, highlighting the degradation of prior capabilities.

\begin{figure}[htbp]
    \centering
    \includegraphics[width=\linewidth]{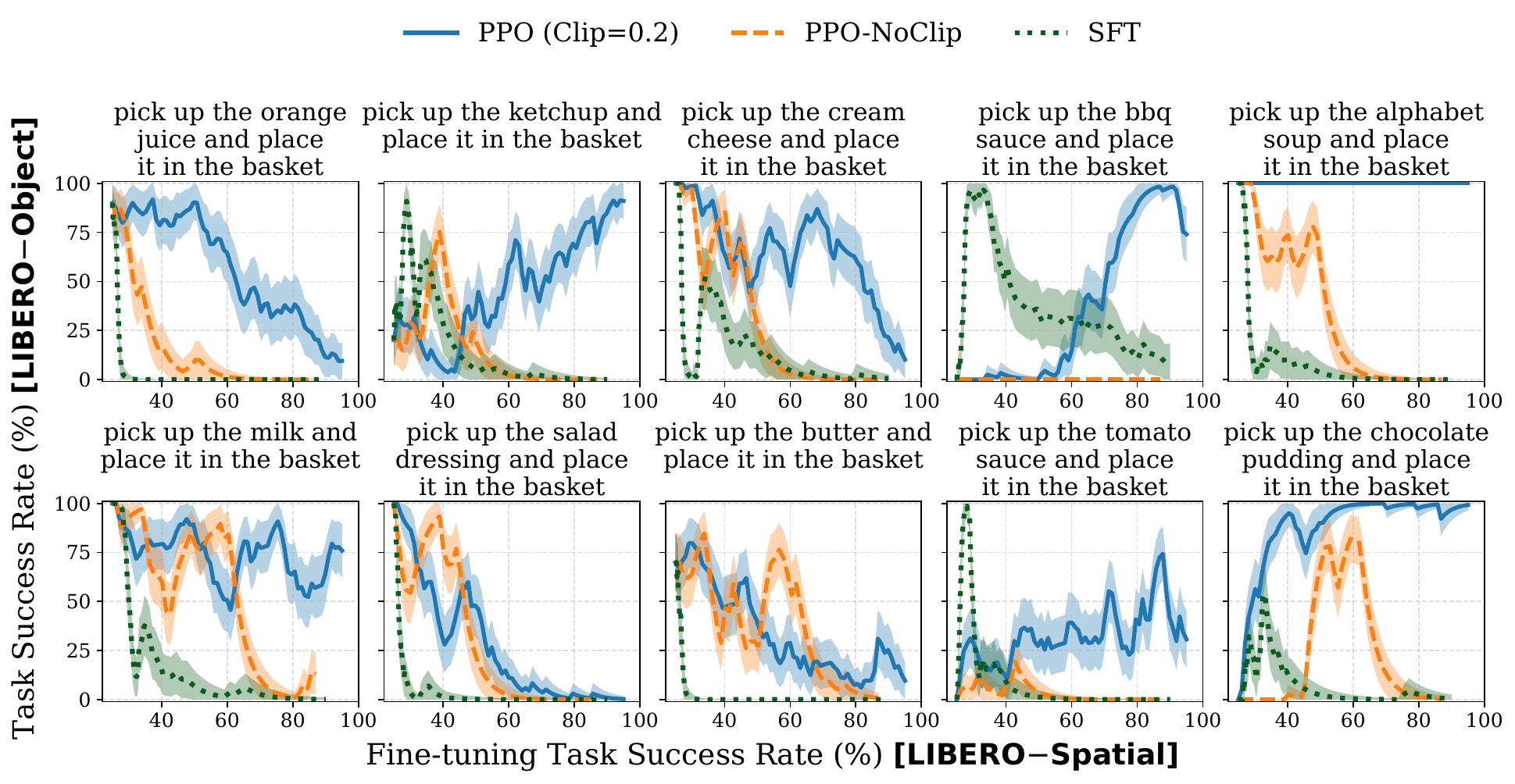}
    \caption{\textbf{Per-task capability retention on the LIBERO-Object suite.} Performance evolution on the held-out Object tasks during downstream adaptation to the Spatial target. Unconstrained methods exhibit rapid performance degradation, whereas trust-region bounds delay this decline.}
    \label{fig:app_object_curves}
\end{figure}

Figure~\ref{fig:app_object_curves} details the retention trajectories across all 10 tasks in the \textbf{LIBERO-Object} suite. As the policy converges on the spatial target, unconstrained optimization (vanilla SFT and PPO-NoClip) drives a rapid drop in performance across diverse manipulation tasks (e.g., picking up orange juice or ketchup). In contrast, bounding the parameter updates via the PPO trust region ($\epsilon=0.2$) delays this degradation.

\begin{figure}[htbp]
    \centering
    \includegraphics[width=\linewidth]{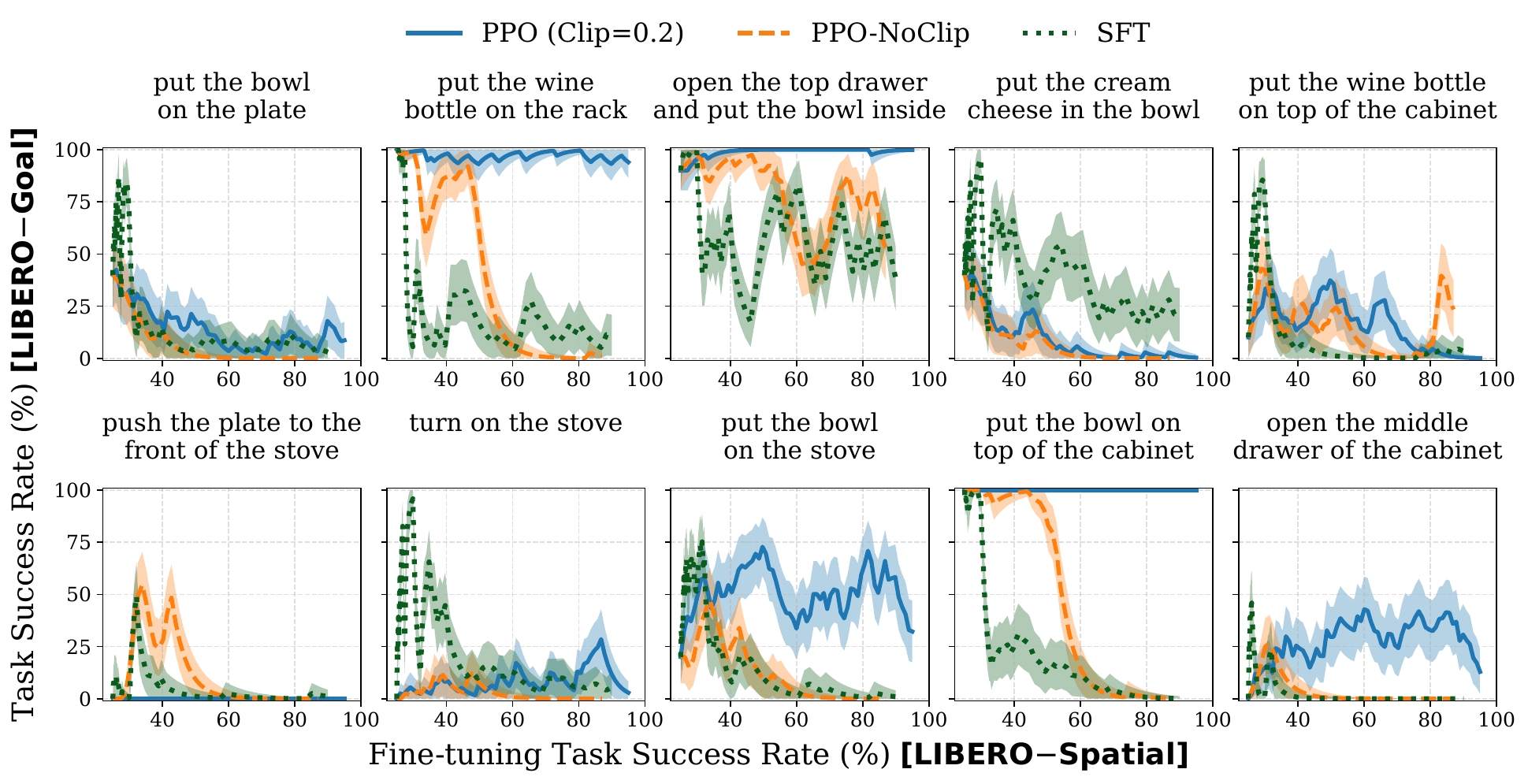}
    \caption{\textbf{Per-task capability retention on the LIBERO-Goal suite.} Performance evolution on the held-out Goal tasks. The trajectories confirm that unconstrained fine-tuning degrades pre-trained sequential execution capabilities.}
    \label{fig:app_goal_curves}
\end{figure}

Figure~\ref{fig:app_goal_curves} presents the analogous evaluation for the 10 tasks in the \textbf{LIBERO-Goal} suite, which involves long-horizon semantic goals (e.g., "open the top drawer" or "put the bowl on the plate"). Consistent with the Object suite dynamics, the absence of trust-region constraints degrades pre-trained sequential behaviors.

\subsection{Formal quantification of update sparsity}
\label{app:subsec:sparsity_definition}

Analyses in Sections~\ref{subsec:3_sparsity_analysis} and~\ref{subsec:5_sparsity_analysis} demonstrate that bounded optimization algorithms, whether via RL trust regions or conservative weighting, correlate with high parameter update sparsity. These bounding mechanisms restrict modifications to a small subnetwork while preserving foundational parameters \cite{mukherjee2026rlsubnetworks}. To quantify this sparsity, we account for the continuous nature of gradient-based optimization, which rarely produces exact zero-magnitude updates.

We define sparsity using a relative deviation threshold. Let $\Theta_{\mathrm{pre}}$ and $\Theta_{\mathrm{ft}}$ denote the flattened parameter vectors of the pre-trained and fine-tuned models, respectively. For a network comprising $N$ parameters, the global update sparsity $S$ is computed as:
\begin{equation}
S = \frac{1}{N} \sum_{i=1}^{N} \mathbb{I} \left( \frac{|(\Theta_{\mathrm{ft}})_{i} - (\Theta_{\mathrm{pre}})_{i}|}{|(\Theta_{\mathrm{pre}})_{i}| + \epsilon} < \delta \right),
\end{equation}
where $\mathbb{I}(\cdot)$ denotes the indicator function, $\epsilon = 10^{-8}$ ensures numerical stability, and $\delta$ represents the relative displacement threshold. 

Across all sparsity evaluations (including Figures~\ref{fig:sparsity} and~\ref{fig:sparsity_heatmap}), we set $\delta = 10^{-3}$, classifying a parameter as unchanged if its relative shift remains below $0.1\%$. This metric distinguishes the sparse parameter updates of bounded methods from the dense global overwrites of vanilla SFT.

%% file: Appendix/2_appendix_method.tex
\section{Extended theoretical analysis: robustness and risk attenuation}
\label{app:sec:theoretical_analysis}

This section provides supplementary theoretical analyses supporting the formulation of the ConSFT objective in Section \ref{sec:methodology}. We address the algorithmic robustness to demonstration noise (Appendix \ref{app:subsec:error_floor}) and formally prove the parameter risk attenuation without relying on prior data (Appendix \ref{app:subsec:fim_decoupling}).

\subsection{Constant probability assumption for human actions}
\label{app:subsec:error_floor}

In Section \ref{subsec:conservative_weighting}, we conceptually assume an ideal expert distribution where the baseline flow matching loss evaluates to strictly zero ($\mathcal{L}_{\mathrm{behavior}} = 0$). In practical physical systems, human demonstrations inherently exhibit epistemic uncertainty and multimodal variance, imposing an irreducible error floor ($\mathcal{L}_{\mathrm{behavior}} = \epsilon_{\mathrm{min}} > 0$). Substituting this physical lower bound into the importance weight formulation factors out a strictly positive global constant multiplier across all samples. Because a global scalar multiplier does not alter relative gradient directions or the proportional weighting between different transitions within a batch, it exerts no influence on the underlying optimization dynamics. Consequently, the presence of demonstration noise and the associated error floor do not compromise the theoretical formulation or the practical implementation of the ConSFT objective.

\subsection{Analytical bounds on relative risk attenuation}
\label{app:subsec:fim_decoupling}

In Section \ref{subsec:risk_analysis}, we utilize the Fisher Information Matrix ($F$) to mathematically quantify the forgetting risk. Because evaluating the exact entries of $F$ strictly requires access to inaccessible pre-training data, $F$ serves as an analytical framework rather than a computable algorithmic component.

\textbf{Relative risk attenuation via quadratic scaling.} 
Because $F$ is inherently positive semi-definite (PSD), the forgetting risk is strictly non-negative: $\mathcal{R}(g) = g^{\top} F g \ge 0$. For any parameter gradient $g$ and a scalar modulation weight $\omega \in (0, 1]$, the quadratic form dictates that the risk scales strictly by $\omega^2$:
\begin{equation}
    \mathcal{R}(\omega \cdot g) = (\omega \cdot g)^{\top} F (\omega \cdot g) = \omega^2 \cdot \mathcal{R}(g).
\end{equation}
In ConSFT, the dynamic weight $\omega = \exp(-\mathcal{L}_{\mathrm{SFT}}(\theta)/\tau)$ is computed solely from the current target-task regression loss. Substituting this scalar yields the relative risk attenuation:
\begin{equation}
    \mathcal{R}(g_{\mathrm{ConSFT}}) = \exp\left(-\frac{2\mathcal{L}_{\mathrm{SFT}}(\theta)}{\tau}\right) \mathcal{R}(g_{\mathrm{SFT}}).
\end{equation}
Since $\exp(-2\mathcal{L}_{\mathrm{SFT}}(\theta)/\tau) \le 1$, this scaling mechanism guarantees that for any high-loss target transition, the forgetting risk attenuates exponentially toward zero relative to the unconstrained gradient $g_{\mathrm{SFT}}$. Crucially, this relative attenuation holds analytically across any arbitrary curvature $F$. By relying on scalar gradient modulation rather than explicit evaluations of the parameter-space geometry, ConSFT structurally suppresses the forgetting risk while strictly adhering to a zero-prior-data regime.

%% file: Appendix/3_appendix_experiments.tex
\section{Benchmark evaluations: VLA fine-tuning and physical deployment}
\label{app:sec:implementation_and_results}

\subsection{Temperature annealing mechanism}
\label{app:subsec:temperature_schedule}

During the early phases of fine-tuning, gradients from unfamiliar target distributions carry a high disruption risk. To bound parameter deviations during this phase while enabling target task convergence, ConSFT employs a dynamic temperature schedule. 

We monotonically increase the temperature $\tau$ over training, annealing the conservative weight $\omega(\theta) = \max\left(\omega_{\mathrm{min}}, \exp\left(-\kappa \frac{\mathcal{L}_{\mathrm{SFT}}(\theta)}{\tau}\right)\right)$ toward $1.0$ to eventually recover the unconstrained SFT objective. The scaling factor $\kappa$ calibrates the penalty on high-loss transitions. Empirically, because physical demonstrations exhibit heavy-tailed multimodal variance, a strictly unbounded exponential decay may entirely suppress valid out-of-distribution behaviors. The lower bound $\omega_{\mathrm{min}}$ acts as a numerical safeguard, ensuring a minimum gradient flow for these high-variance samples to avert gradient vanishing without compromising the overall bounded optimization dynamic. Let $t$ denote the current training step and $T$ the total scheduling steps. We define a normalized decay factor $\rho(t) \in [1]$ as:
\begin{equation}
\rho(t) = \frac{\exp\left(-\lambda \left[\min\left(\frac{t}{T}, 1\right)\right]^c \right) - \exp(-\lambda)}{1 - \exp(-\lambda)},
\end{equation}

where $\lambda$ governs the exponential decay rate and $c$ is the curvature factor. Setting $c > 1.0$ ensures the relaxation is gradual during early training to preserve prior capabilities when the network is most susceptible to forgetting, before accelerating later in the optimization process.

The dynamic temperature $\tau(t)$ is computed as:
\begin{equation}
\tau(t) = \min \left( \frac{\tau_{\mathrm{start}}}{\max(\rho(t), \epsilon)}, \tau_{\mathrm{end}} \right),
\end{equation}
where $\tau_{\mathrm{start}}$ is the initial temperature and $\epsilon = 10^{-8}$ ensures numerical stability. In our VLA experiments, we configure this schedule with $\tau_{\mathrm{start}} = 0.003$, $\lambda = 0.8$, $c = 3.5$, $\kappa = 25.0$, $\omega_{\mathrm{min}} = 0.001$, and $T = 2000$ steps. This schedule provides a controlled transition from conservative parameter anchoring to unconstrained regression.

\textbf{Hyperparameter robustness.} Although the temperature schedule introduces multiple variables, it does not expand the hyperparameter search space. Empirically, the shape-controlling variables ($\lambda$, $c$, $\kappa$) and the boundary constant ($\omega_{\mathrm{min}}$) function as task-agnostic constants. Across all evaluations in this study, these four parameters remain fixed. Only the initial temperature $\tau_{\mathrm{start}}$ (dictating the initial conservatism) and the annealing duration $T$ (scaling with dataset capacity) require task-specific tuning.

\subsection{Optimization and distributed training configuration}
\label{app:subsec:vla_hyperparameters}

Table \ref{tab:vla_hyperparameters} lists the hyperparameters used for the downstream fine-tuning tasks. To manage the memory requirements of the billion-parameter VLA backbone during continuous vector field prediction, we employ a Fully Sharded Data Parallel (FSDP) strategy across 8 NVIDIA A100 (80GB) GPUs. This distributed infrastructure shards model parameters, optimizer states, and gradients, enabling a global batch size of 1024 during the computation of the conservative objective.

\begin{table}[ht]
\caption{\textbf{Hyperparameter configuration for VLA fine-tuning.} Optimization parameters, hardware distribution, and ConSFT temperature scheduling details.}
\label{tab:vla_hyperparameters}
\centering
\begin{tabular}{lc}
\toprule
\textbf{Hyperparameter} & \textbf{Value} \\
\midrule
\multicolumn{2}{l}{\textit{Hardware \& Batching}} \\
Hardware Distribution & 8$\times$ NVIDIA A100 (80GB) with FSDP \\
Global Batch Size & 1024 \\
Micro-batch Size (per GPU) & 32 \\
\midrule
\multicolumn{2}{l}{\textit{Optimization \& Regularization}} \\
Optimizer & Adam \\
Learning Rate & $2.5 \times 10^{-5}$ \\
Adam Betas ($\beta_1, \beta_2$) & (0.9, 0.95) \\
Adam Epsilon ($\epsilon$) & $1.0 \times 10^{-8}$ \\
Weight Decay & $1.0 \times 10^{-10}$ \\
Max Gradient Norm & 1.0 \\
\midrule
\multicolumn{2}{l}{\textit{ConSFT Temperature Scheduling}} \\
Initial Temperature ($\tau_{\mathrm{start}}$) & 0.003 \\
Terminal Temperature ($\tau_{\mathrm{end}}$) & 5.0 \\
Curvature Factor ($c$) & 3.5 \\
Decay Rate ($\lambda$) & 0.8 \\
Scaling Factor ($\kappa$) & 25.0 \\
Lower Bound ($\omega_{\mathrm{min}}$) & 0.001 \\
Decay Steps ($T$) & 2000 \\
\bottomrule
\end{tabular}
\end{table}

\subsection{Baseline adaptations for flow-matching policies}
\label{app:subsec:baseline_details}

This subsection details the configuration of the anti-forgetting baselines evaluated in Section \ref{subsec:baseline_comparison}. These baselines are adapted for the continuous flow-matching architecture of the VLA models.

\textbf{KL-Regularization (LwF).} For the Learning without Forgetting (LwF) \cite{li2017learning} baseline, we instantiate a frozen copy of the pre-trained base model as the reference policy $\pi_{\mathrm{ref}}$. During training on the target task, we apply a regularization penalty to align the outputs of the training policy $\pi_{\theta}$ and the reference policy $\pi_{\mathrm{ref}}$. Because flow-matching networks predict continuous vector fields rather than tractable discrete probabilities, calculating the exact Kullback-Leibler (KL) divergence of the implicit action distribution is intractable. We therefore realize this KL penalty mathematically via the Mean Squared Error (MSE) between the predicted vector fields across sampled denoising timesteps $\tau$. The objective is formulated as:
\begin{equation}
\mathcal{L}_{\mathrm{LwF}}(\theta) = \mathcal{L}_{\mathrm{SFT}}(\theta) + \lambda_{\mathrm{LwF}} \mathbb{E}_{\tau} \left[ \left\| \pi_{\theta}(\cdot \mid \tau) - \pi_{\mathrm{ref}}(\cdot \mid \tau) \right\|_2^2 \right],
\end{equation}
where $\mathcal{L}_{\mathrm{SFT}}(\theta)$ represents the standard flow-matching loss. The scalar $\lambda_{\mathrm{LwF}}$ controls the regularization strength. To stabilize optimization during early training, we set $\lambda_{\mathrm{LwF}} = 0.01$ across all LwF experiments.

\textbf{Experience Replay (ER).} The Experience Replay (ER) \cite{rolnick2019experience} baseline mitigates forgetting by rehearsing historical trajectories. We construct a prior data buffer $\mathcal{B}_{\mathrm{prior}}$ by sampling uniformly from the pre-training datasets. During fine-tuning, each training batch is constructed by sampling examples from the target task dataset $\mathcal{D}_{\mathrm{target}}$ and the prior buffer $\mathcal{B}_{\mathrm{prior}}$ at a 1:1 ratio ($50\%$ target, $50\%$ prior). The model is optimized using the standard flow-matching objective over these mixed batches to balance target task adaptation with foundational skill retention.

\textbf{Low-Rank Adaptation (LoRA).} To evaluate structural constraints as a forgetting mitigation mechanism \cite{biderman2024lora, bohnet2025comparative}, we freeze the pre-trained base model and inject trainable adapter matrices into the linear layers. We configure all adapters with a rank of $r=16$, strictly confining the flow-matching parameter updates to low-rank subspaces.

\subsection{Detailed capability retention trajectories across optimization paradigms}
\label{app:subsec:consft_per_task_success}

\begin{figure}[htbp]
    \centering
    \includegraphics[width=\linewidth]{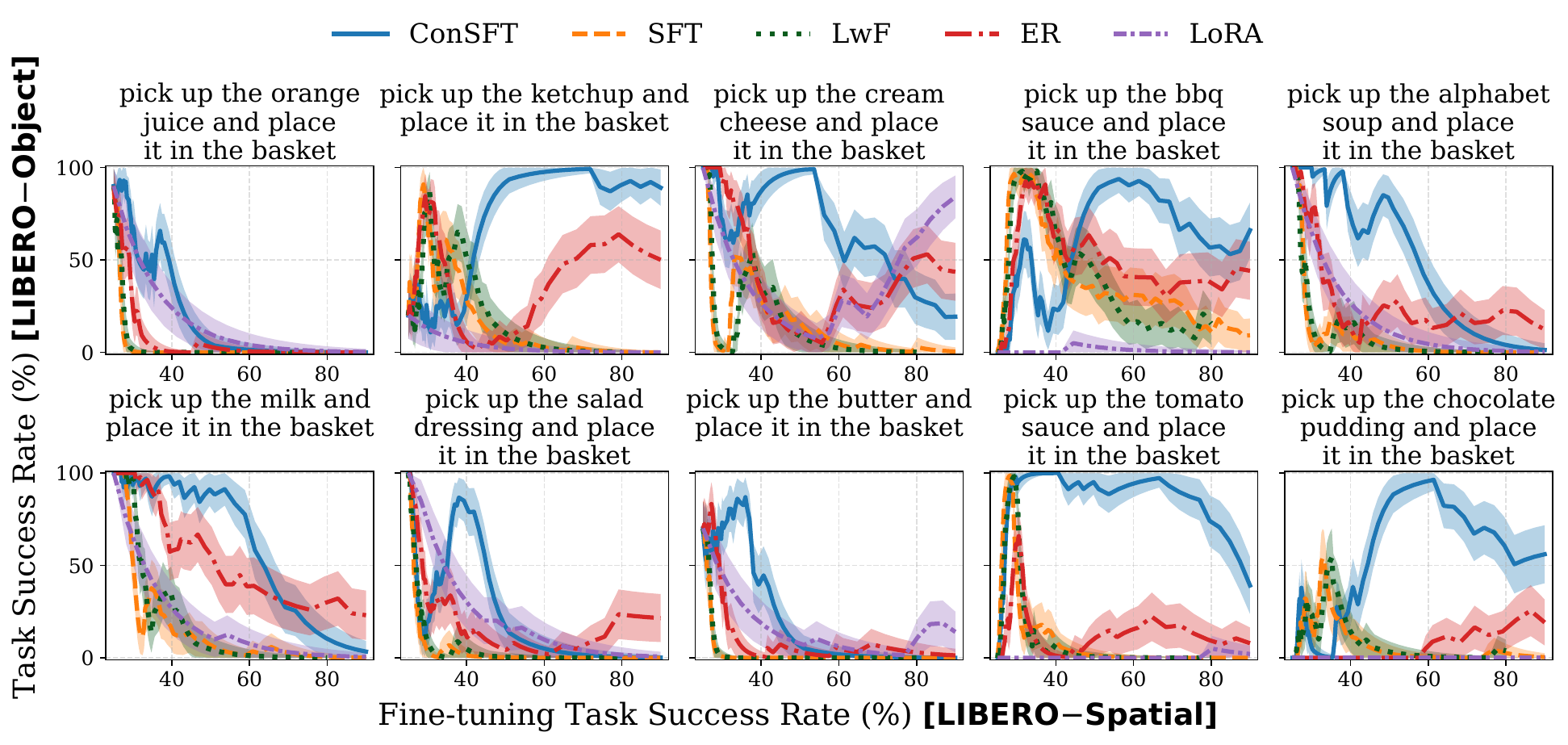}
    \caption{\textbf{Per-task capability retention on the LIBERO-Object suite.} Performance evolution on the held-out Object tasks during adaptation to the Spatial target task.}
    \label{fig:app_object_consft}
\end{figure}

\begin{figure}[htbp]
    \centering
    \includegraphics[width=\linewidth]{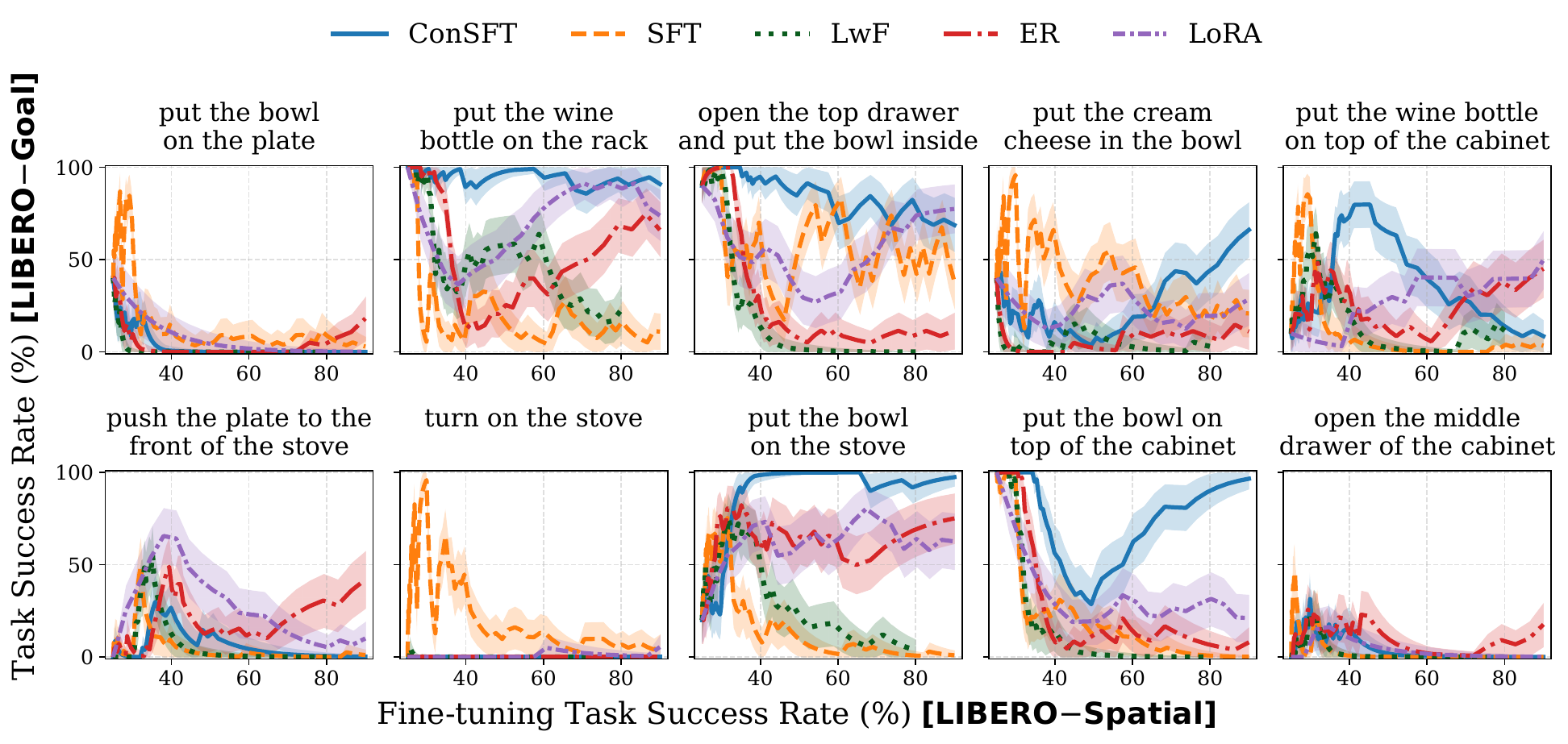}
    \caption{\textbf{Per-task capability retention on the LIBERO-Goal suite.} Performance evolution on the held-out Goal tasks. The multi-baseline trajectories demonstrate that ConSFT bounds the disruption risk, preserving pre-trained long-horizon behaviors in a prior-data-free regime.}
    \label{fig:app_goal_consft}
\end{figure}

To examine capability retention at the task level, we track success rate trajectories during downstream adaptation to \textit{LIBERO-Spatial Task 9}. Figures \ref{fig:app_object_consft} and \ref{fig:app_goal_consft} detail the performance evolution across the \textbf{LIBERO-Object} and \textbf{LIBERO-Goal} suites, comparing ConSFT against vanilla SFT and the anti-forgetting baselines (LwF, ER).

These trajectories illustrate distinct optimization dynamics. Unconstrained vanilla SFT drives rapid performance degradation on prior tasks. KL-Regularization (LwF) exhibits optimization rigidity, delaying but ultimately failing to prevent the loss of pre-trained capabilities. While Experience Replay (ER) maintains foundational skills via explicit data rehearsal, ConSFT bounds capability degradation directly through conservative weighting. By confining target-task updates, ConSFT preserves foundational behaviors without requiring prior data buffers. This empirical stabilization demonstrates that the conservative objective effectively isolates new skill acquisition from foundational parameter disruption.

\subsection{Real-world deployment: foundational robustness and high-precision adaptation}
\label{app:subsec:real_world_deployment}

To establish the evaluation baseline under physical hardware constraints and sensory noise, we deploy the pre-trained $\pi_{0.5}$ backbone on a real-world single-arm robotic system. This protocol evaluates the zero-shot robustness of foundational capabilities and the skill preservation of the fine-tuned model during downstream execution.

\begin{figure}[htbp]
    \centering
    \includegraphics[width=\linewidth]{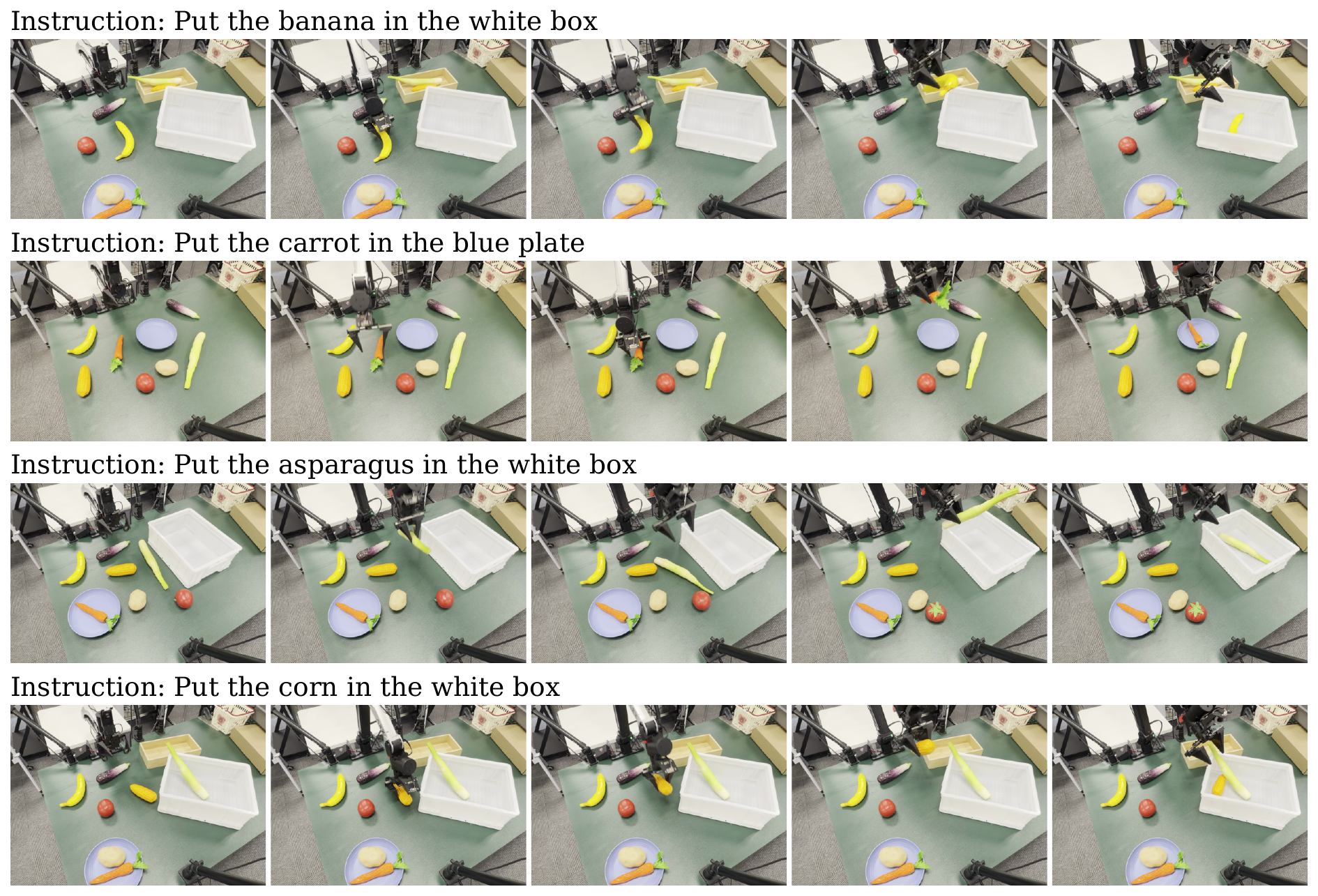}
    \caption{\textbf{Real-world multi-task evaluation under visually dense conditions.} Execution trajectories of the pre-trained $\pi_{0.5}$ policy across four distinct semantic grasping tasks. The environment introduces physical distractors to test foundational robustness prior to downstream adaptation.}
    \label{fig:app_four_tasks_grid}
\end{figure}

\textbf{Foundational robustness in physical environments.} To establish an empirical baseline, we first evaluate the pre-trained $\pi_{0.5}$ policy across multiple semantic grasping tasks. As illustrated in Figure \ref{fig:app_four_tasks_grid}, the workspace introduces physical distractors. The pre-trained backbone exhibits robust initial capabilities, achieving success rates exceeding 80\% in clutter-free scenarios and maintaining success rates above 60\% in visually dense conditions.

\begin{figure}[htbp]
    \centering
    \includegraphics[width=\linewidth]{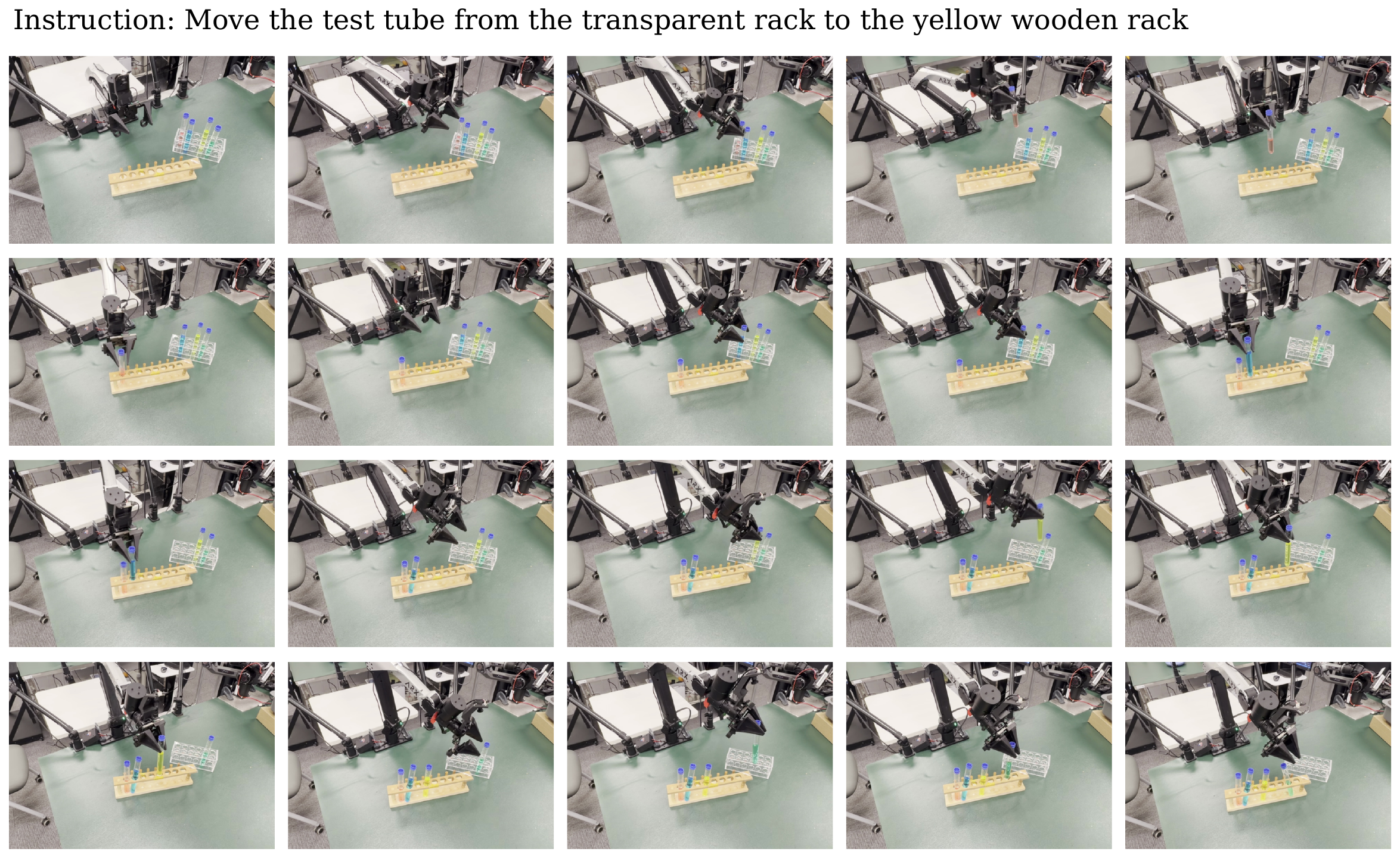}
    \caption{\textbf{Real-world execution of the sequential test tube transfer task.} The single-arm robotic system operates under the ConSFT-optimized policy. The task demands high-precision insertion and long-horizon planning, requiring the sequential transfer of all four test tubes to satisfy the binary success criterion.}
    \label{fig:app_real_world_tube}
\end{figure}

\textbf{High-precision adaptation and capability retention.} Subsequently, we evaluate the ConSFT-optimized policy on a high-precision, long-horizon task: sequential test tube transfer and insertion (Figure \ref{fig:app_real_world_tube}). This target task demands fine-grained continuous control for the single-arm system to grasp, transfer, and insert four distinct test tubes into a specific receptacle.

Across both evaluation paradigms, we enforce a binary success criterion to prevent metric inflation from partial completions. For the semantic grasping tasks, the target object must be released entirely within the geometric boundaries of the designated container. For the sequential insertion task, an episode is recorded as successful if and only if all four test tubes are successfully inserted. Under these criteria, the ConSFT-optimized policy achieves a 70\% success rate on the long-horizon insertion task. This empirical result confirms that ConSFT scales effectively to physical environments, acquiring high-precision target skills while bounding the disruption of foundational parameters.

%% file: references.bib
@article{shenfeld2025rlsrazor,
  title={Rl's razor: Why online reinforcement learning forgets less},
  author={Shenfeld, Idan and Pari, Jyothish and Agrawal, Pulkit},
  journal={arXiv preprint arXiv:2509.04259},
  year={2025}
}

@article{mukherjee2026rlsubnetworks,
  title={Reinforcement learning finetunes small subnetworks in large language models},
  author={Mukherjee, Sagnik and Yuan, Lifan and Hakkani-Tur, Dilek and Peng, Hao},
  journal={Advances in Neural Information Processing Systems},
  volume={38},
  pages={132119--132138},
  year={2026}
}

@article{luo2025empirical,
  title={An empirical study of catastrophic forgetting in large language models during continual fine-tuning},
  author={Luo, Yun and Yang, Zhen and Meng, Fandong and Li, Yafu and Zhou, Jie and Zhang, Yue},
  journal={IEEE Transactions on Audio, Speech and Language Processing},
  year={2025},
  publisher={IEEE}
}

@article{imanov2026mechanistic,
  title={Mechanistic Analysis of Catastrophic Forgetting in Large Language Models During Continual Fine-tuning},
  author={Imanov, Olaf Yunus Laitinen},
  journal={arXiv preprint arXiv:2601.18699},
  year={2026}
}

@article{chu2025sft,
  title={Sft memorizes, rl generalizes: A comparative study of foundation model post-training},
  author={Chu, Tianzhe and Zhai, Yuexiang and Yang, Jihan and Tong, Shengbang and Xie, Saining and Schuurmans, Dale and Le, Quoc V and Levine, Sergey and Ma, Yi},
  journal={arXiv preprint arXiv:2501.17161},
  year={2025}
}

@article{biderman2024lora,
  title={Lo{RA} Learns Less and Forgets Less},
  author={Dan Biderman and Jacob Portes and Jose Javier Gonzalez Ortiz and Mansheej Paul and Philip Greengard and Connor Jennings and Daniel King and Sam Havens and Vitaliy Chiley and Jonathan Frankle and Cody Blakeney and John Patrick Cunningham},
  journal={Transactions on Machine Learning Research},
  issn={2835-8856},
  year={2024},
  url={https://openreview.net/forum?id=aloEru2qCG},
  note={Featured Certification}
}

@article{bohnet2025comparative,
  title={A Comparative Analysis of LLM Adaptation: SFT, LoRA, and ICL in Data-Scarce Scenarios},
  author={Bohnet, Bernd and Dangovski, Rumen and Swersky, Kevin and Moore, Sherry and Chaudhry, Arslan and Kenealy, Kathleen and Fiedel, Noah},
  journal={arXiv preprint arXiv:2511.00130},
  year={2025}
}

@article{sun2026safety,
  title={Safety alignment as continual learning: Mitigating the alignment tax via orthogonal gradient projection},
  author={Sun, Guanglong and Zhang, Siyuan and Wang, Liyuan and Zhu, Jun and Su, Hang and Zhong, Yi},
  journal={arXiv preprint arXiv:2602.07892},
  year={2026}
}

@article{phan2025beyond,
  title={Beyond Reasoning Gains: Mitigating General Capabilities Forgetting in Large Reasoning Models},
  author={Phan, Hoang and Yang, Xianjun and Yao, Kevin and Zhang, Jingyu and Bi, Shengjie and Tang, Xiaocheng and Khabsa, Madian and Liu, Lijuan and Lei, Deren},
  journal={arXiv preprint arXiv:2510.21978},
  year={2025}
}

@article{mcallister2025fpo,
  title={Flow Matching Policy Gradients},
  author={McAllister, David and Ge, Songwei and Yi, Brent and Kim, Chung Min and Weber, Ethan and Choi, Hongsuk and Feng, Haiwen and Kanazawa, Angjoo},
  journal={arXiv preprint arXiv:2507.21053},
  year={2025}
}

@article{lai2025rft,
  title={Reinforcement Fine-Tuning Naturally Mitigates Forgetting in Continual Post-Training},
  author={Lai, Song and Zhao, Haohan and Feng, Rong and Ma, Changyi and Liu, Wenzhuo and Zhao, Hongbo and Lin, Xi and Yi, Dong and Xie, Min and Zhang, Qingfu and Liu, Hongbin and Meng, Gaofeng and Zhu, Fei},
  journal={arXiv preprint arXiv:2507.05386},
  year={2025}
}

@article{hu2026simplerecipe,
  title={Simple Recipe Works: Vision-Language-Action Models are Natural Continual Learners with Reinforcement Learning},
  author={Hu, Jiaheng and Shim, Jay and Tang, Chen and Sung, Yoonchang and Liu, Bo and Stone, Peter and Martin-Martin, Roberto},
  journal={arXiv preprint arXiv:2603.11653},
  year={2026}
}

@article{liu2026pretrained,
  title={Pretrained Vision-Language-Action Models are Surprisingly Resistant to Forgetting in Continual Learning},
  author={Liu, Huihan and Kim, Changyeon and Liu, Bo and Liu, Minghuan and Zhu, Yuke},
  journal={arXiv preprint arXiv:2603.03818},
  year={2026}
}

@article{liu2026towards,
  title={Towards Long-Lived Robots: Continual Learning VLA Models via Reinforcement Fine-Tuning},
  author={Liu, Yuan and Li, Haoran and Tian, Shuai and Qin, Yuxing and Chen, Yuhui and Zheng, Yupeng and Huang, Yongzhen and Zhao, Dongbin},
  journal={arXiv preprint arXiv:2602.10503},
  year={2026}
}

@article{hancock2025actions,
  title={Actions as language: Fine-tuning vlms into vlas without catastrophic forgetting},
  author={Hancock, Asher J and Wu, Xindi and Zha, Lihan and Russakovsky, Olga and Majumdar, Anirudha},
  journal={arXiv preprint arXiv:2509.22195},
  year={2025}
}

@article{black2024pi0,
  title={$\pi_0$: A Vision-Language-Action Flow Model for General Robot Control},
  author={Black, Kevin and Brown, Noah and Driess, Danny and Esmail, Adnan and Equi, Michael and Finn, Chelsea and Fusai, Niccolo and Groom, Lachy and Hausman, Karol and Ichter, Brian and Jakubczak, Szymon and Jones, Tim and Ke, Liyiming and Levine, Sergey and Li-Bell, Adrian and Mothukuri, Mohith and Nair, Suraj and Pertsch, Karl and Shi, Lucy Xiaoyang and Tanner, James and Vuong, Quan and Walling, Anna and Wang, Haohuan and Zhilinsky, Ury},
  journal={arXiv preprint arXiv:2410.24164},
  year={2024}
}

@inproceedings{black2025pi05,
  title={$\pi_{0.5}$: A Vision-Language-Action Model with Open-World Generalization},
  author={Black, Kevin and Brown, Noah and Darpinian, James and Dhabalia, Karan and Driess, Danny and Esmail, Adnan and Equi, Michael Robert and Finn, Chelsea and Fusai, Niccolo and Galliker, Manuel Y. and Ghosh, Dibya and Groom, Lachy and Hausman, Karol and Ichter, Brian and Jakubczak, Szymon and Jones, Tim and Ke, Liyiming and LeBlanc, Devin and Levine, Sergey and Li-Bell, Adrian and Mothukuri, Mohith and Nair, Suraj and Pertsch, Karl and Ren, Allen Z. and Shi, Lucy Xiaoyang and Smith, Laura and Springenberg, Jost Tobias and Stachowicz, Kyle and Tanner, James and Vuong, Quan and Walke, Homer and Walling, Anna and Wang, Haohuan and Yu, Lili and Zhilinsky, Ury},
  booktitle={9th Annual Conference on Robot Learning},
  year={2025}
}

@article{bjorck2025groot,
  title={{GR00T N1}: An Open Foundation Model for Generalist Humanoid Robots},
  author={Bjorck, Johan and Casta{\~n}eda, Fernando and Cherniadev, Nikita and Da, Xingye and Ding, Runyu and Fan, Linxi and Fang, Yu and Fox, Dieter and Hu, Fengyuan and Huang, Spencer and Jang, Joel and Jiang, Zhenyu and Kautz, Jan and Kundalia, Kaushil and Lao, Lawrence and Li, Zhiqi and Lin, Zongyu and Lin, Kevin and Liu, Guilin and Llontop, Edith and Magne, Loic and Mandlekar, Ajay and Narayan, Avnish and Nasiriany, Soroush and Reed, Scott and Tan, You Liang and Wang, Guanzhi and Wang, Zu and Wang, Jing and Wang, Qi and Xiang, Jiannan and Xie, Yuqi and Xu, Yinzhen and Xu, Zhenjia and Ye, Seonghyeon and Yu, Zhiding and Zhang, Ao and Zhang, Hao and Zhao, Yizhou and Zheng, Ruijie and Zhu, Yuke},
  journal={arXiv preprint arXiv:2503.14734},
  year={2025}
}

@article{wu2026lingbot,
  title={A Pragmatic {VLA} Foundation Model},
  author={Wu, Wei and Lu, Fan and Wang, Yunnan and Yang, Shuai and Liu, Shi and Wang, Fangjing and Zhu, Qian and Sun, He and Wang, Yong and Ma, Shuailei and Ren, Yiyu and Zhang, Kejia and Yu, Hui and Zhao, Jingmei and Zhou, Shuai and Qiu, Zhenqi and Xiong, Houlong and Wang, Ziyu and Wang, Zechen and Cheng, Ran and Li, Yong-Lu and Huang, Yongtao and Zhu, Xing and Shen, Yujun and Zheng, Kecheng},
  journal={arXiv preprint arXiv:2601.18692},
  year={2026}
}

@article{schulman2017ppo,
  title={Proximal Policy Optimization Algorithms},
  author={Schulman, John and Wolski, Filip and Dhariwal, Prafulla and Radford, Alec and Klimov, Oleg},
  journal={arXiv preprint arXiv:1707.06347},
  year={2017}
}

@inproceedings{schulman2015trpo,
  title={Trust region policy optimization},
  author={Schulman, John and Levine, Sergey and Abbeel, Pieter and Jordan, Michael and Moritz, Philipp},
  booktitle={International conference on machine learning},
  pages={1889--1897},
  year={2015},
  organization={PMLR}
}

@article{sabbaghi2026robust,
  title={Robust Policy Optimization to Prevent Catastrophic Forgetting},
  author={Sabbaghi, Mahdi and Pappas, George and Javanmard, Adel and Hassani, Hamed},
  journal={arXiv preprint arXiv:2602.08813},
  year={2026}
}

@article{gao2025sapo,
  title={Soft adaptive policy optimization},
  author={Gao, Chang and Zheng, Chujie and Chen, Xiong-Hui and Dang, Kai and Liu, Shixuan and Yu, Bowen and Yang, An and Bai, Shuai and Zhou, Jingren and Lin, Junyang},
  journal={arXiv preprint arXiv:2511.20347},
  year={2025}
}

@article{yi2026flow,
  title={Flow Policy Gradients for Robot Control},
  author={Yi, Brent and Choi, Hongsuk and Singh, Himanshu Gaurav and Huang, Xiaoyu and Truong, Takara E and Sferrazza, Carmelo and Ma, Yi and Duan, Rocky and Abbeel, Pieter and Shi, Guanya and Liu, Karen and Kanazawa, Angjoo},
  journal={arXiv preprint arXiv:2602.02481},
  year={2026}
}

@article{zhang2026reinflow,
  title={Reinflow: Fine-tuning flow matching policy with online reinforcement learning},
  author={Zhang, Tonghe and Yu, Chao and Su, Sichang and Wang, Yu},
  journal={Advances in Neural Information Processing Systems},
  volume={38},
  pages={106282--106319},
  year={2026}
}

@article{pfrommer2025reinforcement,
  title={Reinforcement learning for flow-matching policies},
  author={Pfrommer, Samuel and Huang, Yixiao and Sojoudi, Somayeh},
  journal={arXiv preprint arXiv:2507.15073},
  year={2025}
}

@article{kirkpatrick2017overcoming,
  title={Overcoming catastrophic forgetting in neural networks},
  author={Kirkpatrick, James and Pascanu, Razvan and Rabinowitz, Neil and Veness, Joel and Desjardins, Guillaume and Rusu, Andrei A. and Milan, Kieran and Quan, John and Ramalho, Tiago and Grabska-Barwinska, Agnieszka and Hassabis, Demis and Clopath, Claudia and Kumaran, Dharshan and Hadsell, Raia},
  journal={Proceedings of the national academy of sciences},
  volume={114},
  number={13},
  pages={3521--3526},
  year={2017},
  publisher={National Academy of Sciences}
}

@inproceedings{mu2025robotwin,
  title={Robotwin: Dual-arm robot benchmark with generative digital twins},
  author={Mu, Yao and Chen, Tianxing and Chen, Zanxin and Peng, Shijia and Lan, Zhiqian and Gao, Zeyu and Liang, Zhixuan and Yu, Qiaojun and Zou, Yude and Xu, Mingkun and Lin, Lunkai and Xie, Zhiqiang and Ding, Mingyu and Luo, Ping},
  booktitle={Proceedings of the computer vision and pattern recognition conference},
  pages={27649--27660},
  year={2025}
}

@article{liu2023libero,
  title={Libero: Benchmarking knowledge transfer for lifelong robot learning},
  author={Liu, Bo and Zhu, Yifeng and Gao, Chongkai and Feng, Yihao and Liu, Qiang and Zhu, Yuke and Stone, Peter},
  journal={Advances in Neural Information Processing Systems},
  volume={36},
  pages={44776--44791},
  year={2023}
}

@article{rolnick2019experience,
  title={Experience replay for continual learning},
  author={Rolnick, David and Ahuja, Arun and Schwarz, Jonathan and Lillicrap, Timothy and Wayne, Gregory},
  journal={Advances in neural information processing systems},
  volume={32},
  year={2019}
}

@article{li2017learning,
  title={Learning without forgetting},
  author={Li, Zhizhong and Hoiem, Derek},
  journal={IEEE transactions on pattern analysis and machine intelligence},
  volume={40},
  number={12},
  pages={2935--2947},
  year={2017},
  publisher={IEEE}
}

@article{shi2025continual,
  title={Continual learning of large language models: A comprehensive survey},
  author={Shi, Haizhou and Xu, Zihao and Wang, Hengyi and Qin, Weiyi and Wang, Wenyuan and Wang, Yibin and Wang, Zifeng and Ebrahimi, Sayna and Wang, Hao},
  journal={ACM Computing Surveys},
  volume={58},
  number={5},
  pages={1--42},
  year={2025},
  publisher={ACM New York, NY}
}

@article{ouyang2022training,
  title={Training language models to follow instructions with human feedback},
  author={Ouyang, Long and Wu, Jeffrey and Jiang, Xu and Almeida, Diogo and Wainwright, Carroll and Mishkin, Pamela and Zhang, Chong and Agarwal, Sandhini and Slama, Katarina and Ray, Alex and Schulman, John and Hilton, Jacob and Kelton, Fraser and Miller, Luke and Simens, Maddie and Askell, Amanda and Welinder, Peter and Christiano, Paul and Leike, Jan and Lowe, Ryan},
  journal={Advances in neural information processing systems},
  volume={35},
  pages={27730--27744},
  year={2022}
}
